\documentclass[11pt]{article}

\usepackage[final]{acl}

\usepackage{booktabs}
\usepackage[table]{xcolor}
\usepackage{algorithm}
\usepackage{algpseudocode}
\algtext*{EndIf}
\algtext*{EndFor}
\algtext*{EndWhile}
\algtext*{EndFunction}
\usepackage{times}
\usepackage{latexsym}
\usepackage{amsmath}
\usepackage{amsfonts}
\usepackage{amssymb}
\usepackage{hyperref} 
\usepackage{graphicx}
\usepackage{fontawesome5}
\usepackage[T1]{fontenc}

\usepackage[utf8]{inputenc}

\usepackage{microtype}

\usepackage{inconsolata}
\usepackage{multirow}
\usepackage{graphicx}
\usepackage{subcaption}
\algnewcommand{\Input}{\item[\textbf{Input:}]}
\algnewcommand{\Output}{\item[\textbf{Output:}]}
\title{MemDefrag: Latent Memory Defragmentation for Large Language Models}

\author{
  Ruiyi Yan\thanks{This work was done during internship at LIGHTSPEED.} \\
  Kyoto University \\
  \texttt{ruiyi@nlp.ist.i.kyoto-u.ac.jp} \\
  \And
  Zhuoyuan Mao \\
  LIGHTSPEED \\
  \texttt{zhuoyuanmao@global.tencent.com} \\
  \AND
  Yiwen Guo\thanks{Corresponding author.} \\
  Independent Researcher \\
  \texttt{guoyiwen89@gmail.com}
}

\begin{document}
\maketitle
\begin{abstract}
Latent memory, which stores past knowledge fragments as per-layer hidden states, has emerged as a promising paradigm (e.g., MemoryLLM and M+) for long-term memory in large language models (LLMs).
However, the paradigm suffers from significant performance degradation during memory updates, due to positional encoding misalignment and the absence of any tracing mechanism to distinguish target memory fragments from irrelevant ones.
To discover such a tracing mechanism, we probe the layer-wise \textit{attention density} over stored memory fragments, and find that a small set of middle transformer layers consistently concentrates the highest density on the target fragment — exposing an inherent \textit{tracing signal}.
In light of this, we propose \textbf{MemDefrag}, a \textit{training-free} and \textit{model-agnostic} framework that (1) uses a middle-layer tracing signal to conduct memory defragmentation (rank, reorder, and filter memories), and (2) applies an informativeness-guided proportional forgetting mechanism once capacity is exceeded.
Experiments show that MemDefrag substantially outperforms MemoryLLM and M+ on knowledge retention (e.g., 43.0\% vs. 17.4\%/17.6\% after 50 memory updates) and long-context benchmarks, and generalizes well across various LLMs and latent-memory variants. The code is available at \faGithub~\href{https://github.com/ryehr/MemDefrag}{github.com/ryehr/MemDefrag}.
\end{abstract}

\section{Introduction}
\label{sec:introduction}

Despite the impressive capabilities of Large Language Models (LLMs),
a long-standing challenge persists:
\emph{how can a deployed LLM continually absorb new knowledge, while faithfully retaining what has already been stored?}
Previous solutions can be broadly grouped into three classes.
(1)~\textbf{Token-level memory}~\cite{packer2024memgptllmsoperatingsystems, chhikara2025mem0buildingproductionreadyai}
stores past information as explicit text tokens, while it is bottlenecked by the context window and is vulnerable to the \emph{lost-in-the-middle} phenomenon~\cite{liu-etal-2024-lost}.
(2)~\textbf{Parametric memory}
either encodes knowledge into the base model~\cite{zhao2025pretraininglimitedmemorylanguage}
or into auxiliary adapters such as LoRA~\cite{hu2022lora}.
This form incurs no extra inference cost but requires costly
tuning and is susceptible to catastrophic forgetting~\cite{doi:10.1073/pnas.1611835114}.
(3)~\textbf{Latent memory} carries information implicitly in the
model's internal representations such as KV-caches, activations,
or per-layer hidden states. The \textit{long-term} form
(e.g., MemoryLLM~\cite{10.5555/3692070.3694135}, M+~\cite{wang2025m},
and NextMem~\cite{zhang2026nextmemlatentfactualmemory}) maintains a persistent pool of
per-layer hidden states. By avoiding the repeated re-encoding of the full context and compressing information more densely in latent representations than discrete text tokens, this latent paradigm offers a compelling foundation for continually updatable LLMs.

However, the current long-term latent memory paradigm encounters the following limitations. \textbf{(1) Positional encoding misalignment}:
When a new knowledge fragment is appended, the concatenated memory fragments are indexed with positions that no longer match those they were formed at, distorting the attention computation at inference time~\cite{298501}. 
\textbf{(2) The absence of any tracing mechanism}:
Existing methods treat all memory fragments as a flat, undifferentiated prefix and cannot dispel the noise information or single out fragment(s) relevant to the current query.

To address these issues, we propose \textbf{MemDefrag}, a \emph{training-free} and \emph{model-agnostic} framework that reinvents the way long-term latent memory is used and evolved.
MemDefrag is fully \textit{plug-and-play}: it requires neither additional training nor auxiliary modules to the
underlying LLM. Specifically, the contributions of this work are:

(1) We first uncover an inherent \textbf{tracing signal} in transformer layers. Inspired by the query-aware sparsity of tracing token criticality across layer-wise KV cache pages during self-attention~\cite{tang2024quest}, we first ask a diagnostic question:
\emph{is there any intrinsic signal in a transformer, that can trace specific relevant and crucial fragments inside the latent memory?}
By probing the layer-wise \textit{attention density} that each stored fragment receives with respect to a given prompt, we find a clear and robust pattern: a small set of middle transformer layers consistently concentrate the highest attention density on the target fragment.


(2) In light of this finding, we propose inference-time \textbf{memory defragmentation}. At each prompt, MemDefrag computes the attention density of every stored fragment at a \emph{tracer layer} and conducts defragmentation (ranks, reorders, and filters the memory fragments by their density). This simultaneously mitigates the irrelevant semantic noise and the accumulated positional distortion.

(3) Considering a bounded memory and compute budget, we propose an \textbf{informativeness-based} proportional forgetting strategy for memory evolution. When memory capacity is reached, MemDefrag allocates forgetting quotas proportionally across fragments, and within each fragment, evicts the tokens with the \emph{lowest self-information}.

(4) Experiments show that MemDefrag substantially outperforms MemoryLLM and M+ on both knowledge retention and long-context QA.
For instance in knowledge retention, on NaturalQA at step 50, MemDefrag (based on Llama-3.1-8B-Instruct) retains 43.0\% accuracy, compared with 17.4\% and 17.6\% for the two baselines.
Moreover, experiments support the generality and compatibility of our MemDefrag across models, latent-memory variants, and prompt compression.


\section{Investigation: Positional Distortion of Latent Memory}
\label{sec:Investigation}

\begin{figure}[!t]
    \centering
    
    \begin{subfigure}[b]{\linewidth}
        \centering
        \includegraphics[width=\textwidth]{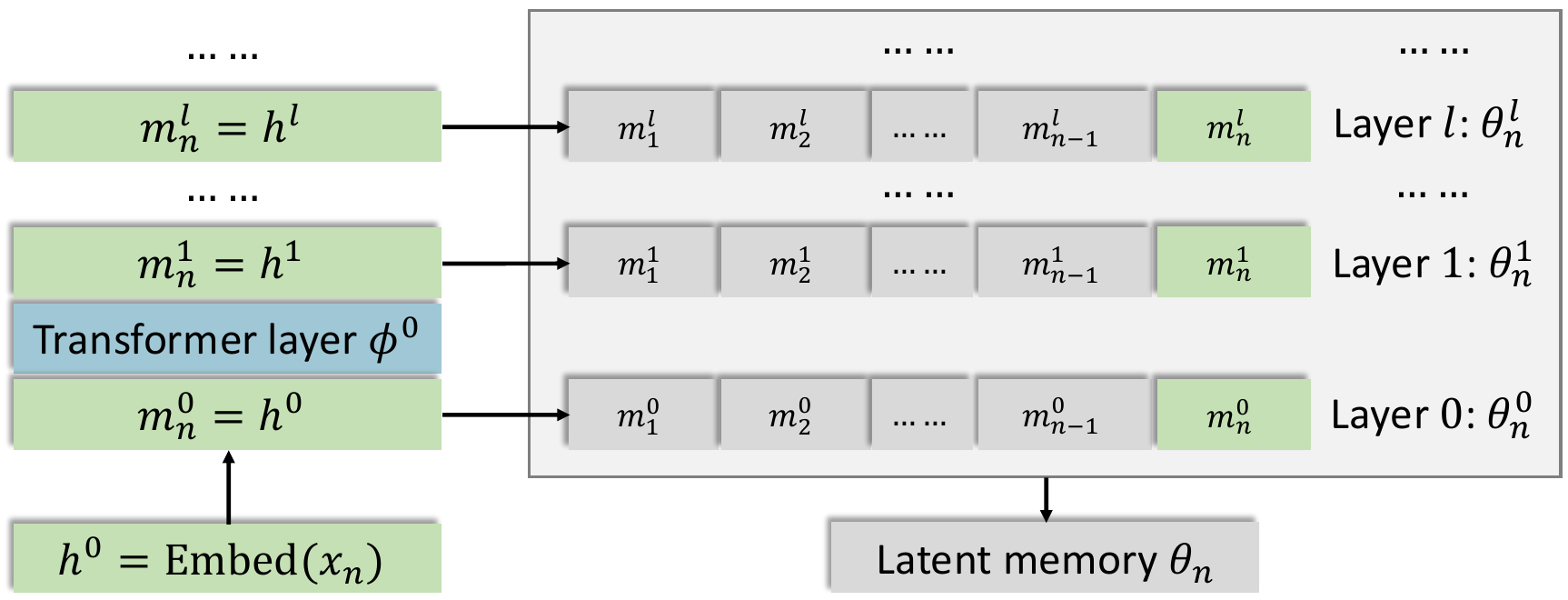}
        \caption{Memory formation of a new knowledge fragment $x_n$ by prefilling in transformer layers. It is concatenated to the existing latent memory $\theta_{n-1}$ to yield $\theta_n$ (memory evolution).}
        \label{fig:vanilla_formation}
    \end{subfigure}
    \hfill
    \begin{subfigure}[b]{\linewidth}
        \centering
        \includegraphics[width=\textwidth]{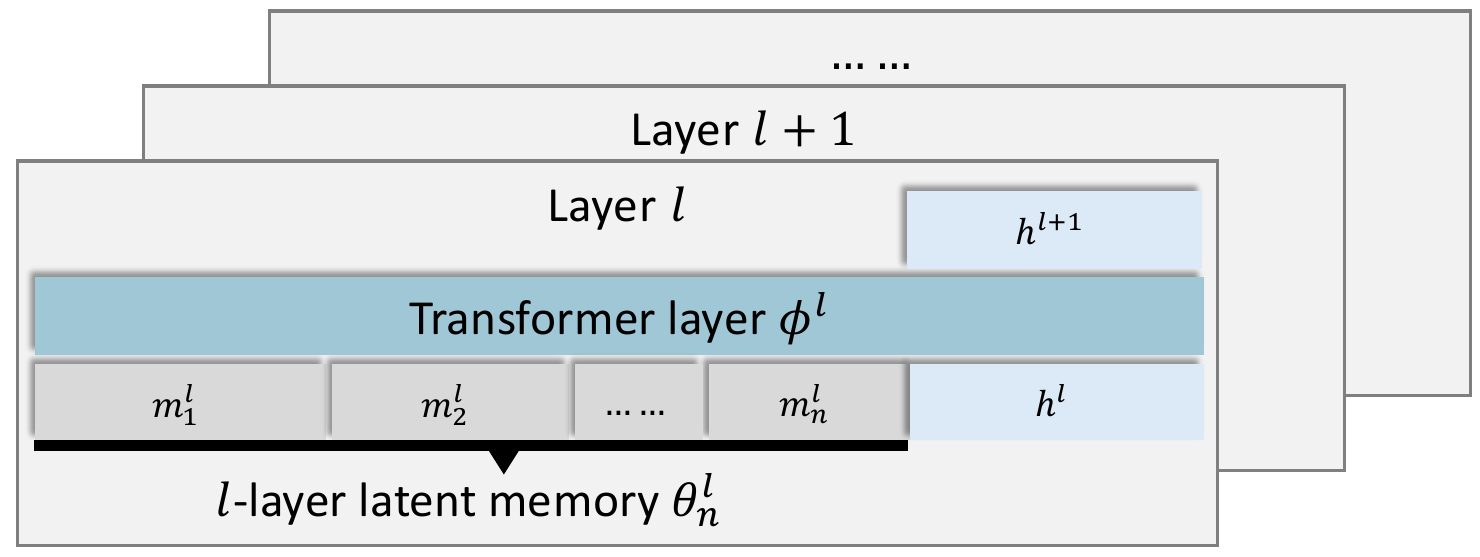}
        \caption{Model inference where the latent memory $\theta_n$ serves as a prefix of hidden states.}
        \label{fig:vanilla_inference}
    \end{subfigure}
  
    \caption{Memory formation, evolution and model inference under the paradigm of vanilla latent memory.}
    \label{fig:vanilla_latent_memory}
\end{figure}

\subsection{Preliminaries: Vanilla Latent Memory}
\label{sec: Vanilla Latent Memory}
To introduce the latent memory paradigm and investigate its positional distortion, we first introduce a \textit{vanilla} framework of latent memory, aiming to exclude irrelevant or redundant designs. 

A language model equipped with latent memory is denoted as $\mathcal{M}_{\theta,\phi}$ comprising two sets of parameters: memory parameters $\theta$ and model parameters $\phi$. 
Specifically, $\phi$ consists of $L$ transformer layers, denoted as $\phi = \{\phi^l\}^L_{l=1}$. The memory parameters $\theta$ are \textbf{prefix} hidden vectors within each transformer layer, denoted as $\theta = \{\theta^l\}^L_{l=1}$. Each $\theta^l$ has dimension $N \times d$, corresponding to $N$ hidden states with hidden dimension $d$ in $\phi$.

Model parameters $\phi$ remain static, while the memory parameters $\theta$ evolve upon encountering new knowledge. Crucially, in a training-free manner, the evolution requires neither gradient descent nor backpropagation.
Given a new textual knowledge fragment as $x$, the evolution is carried out by an update function $U_{\phi}$ (which is based on the model parameters $\phi$), that stores $x$ into $\theta$ to yield updated memory parameters $\theta^\prime$:
\begin{equation}
    \theta^\prime = U_{\phi}(\theta,x).
\end{equation}

Extending this to multi-step, long-term evolution, consider a sequence of textual knowledge fragments $(x_1,\dots,x_n)$, where $x_n$ denotes the knowledge fragment received at time step $n$. The model successively integrates all knowledge to obtain the memory parameters $\theta_{n}$ with its length $N_n$:
\begin{equation}
    \theta_{n} = U_\phi(\cdots(U_\phi(\theta_0,x_1)),x_n).
\end{equation}

\paragraph{Memory Formation by Prefilling}
Consider storing a new knowledge fragment $x_n = [x_n^{(0)}, x_n^{(1)},\dots,x_n^{(|x_n|)}]$, a sequence of tokens at time step $n$. The formed memory fragment $m_n = \{m_n^l\}_{l=1}^{L} = f_{\text{form}}(x_n)$ for $x_n$ consists of the per-layer hidden states produced by passing (prefilling) $x_n$ through the model $\mathcal{M}_{\theta = \emptyset,\phi}$ (i.e., without latent memory):
\begin{equation} 
 \quad m_n^l = \phi_l(h_n^{l-1}) \in \mathbb{R}^{|x_n| \times d}, 
\end{equation}
where $h_n^{l-1}$ denotes the input hidden states to layer $l$, with $h_n^{0}$ being the token embeddings of $x_n$.

\paragraph{Memory Evolution by Concatenation}
Memory parameters are initialized as empty, i.e., $\theta_0 = \emptyset$ (with initial length $N_0 = 0$). Given the current memory parameters $\theta_{n-1}$ and the formed memory fragment $m_n$, the evolution proceeds by concatenation: $\theta_n = \text{concat}(\theta_{n-1}, m_n)$.

\paragraph{Inference with Latent Memory} During inference, at each layer $l$,  latent memory $\theta_n^l$ serves as the prefix of the hidden states $h^l$.

For the whole processes of the paradigm of vanilla latent memory, Figure~\ref{fig:vanilla_latent_memory} shows the sketch of memory formation, evolution and model inference.
MemoryLLM~\cite{10.5555/3692070.3694135}, M+~\cite{wang2025m}, and NextMem~\cite{zhang2026nextmemlatentfactualmemory} represent the latent memory paradigm, and follow the vanilla memory evolution by concatenation and inference with the whole latent memory.

From Figure~\ref{fig:vanilla_latent_memory}, it can be inferred that, in such systems, the position encodings of concatenated memory fragments are misaligned with those expected during inference. 
Note that this positional distortion refers to the position encodings of memory fragments (including the target fragments and other fragments) being misaligned with those expected.
This distortion can be accumulated and significantly degrade generation quality~\cite{298501, li2025seqpetransformersequentialposition}. 
We are therefore motivated to examine the effects of positional distortion.

\begin{figure}[!t]
    \centering
    \begin{subfigure}[b]{0.8\linewidth}
        \centering
        \includegraphics[width=\textwidth]{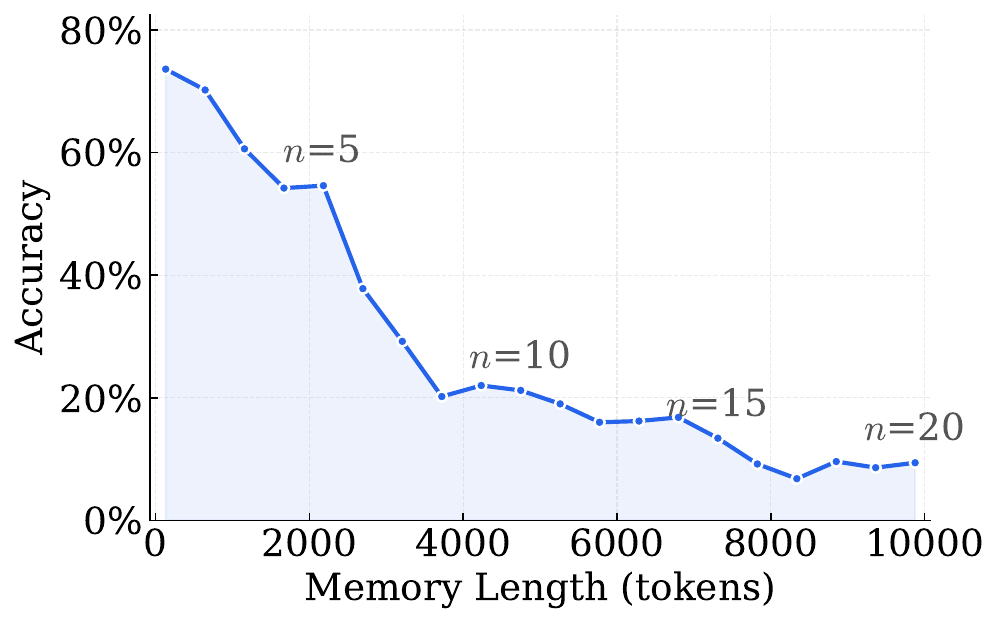}
        \caption{QA accuracy decreasing as the memory length (in tokens) increases.}
        \label{fig:investigation_sub1}
    \end{subfigure}
    \hfill
    \begin{subfigure}[b]{0.8\linewidth}
        \centering
        \includegraphics[width=\textwidth]{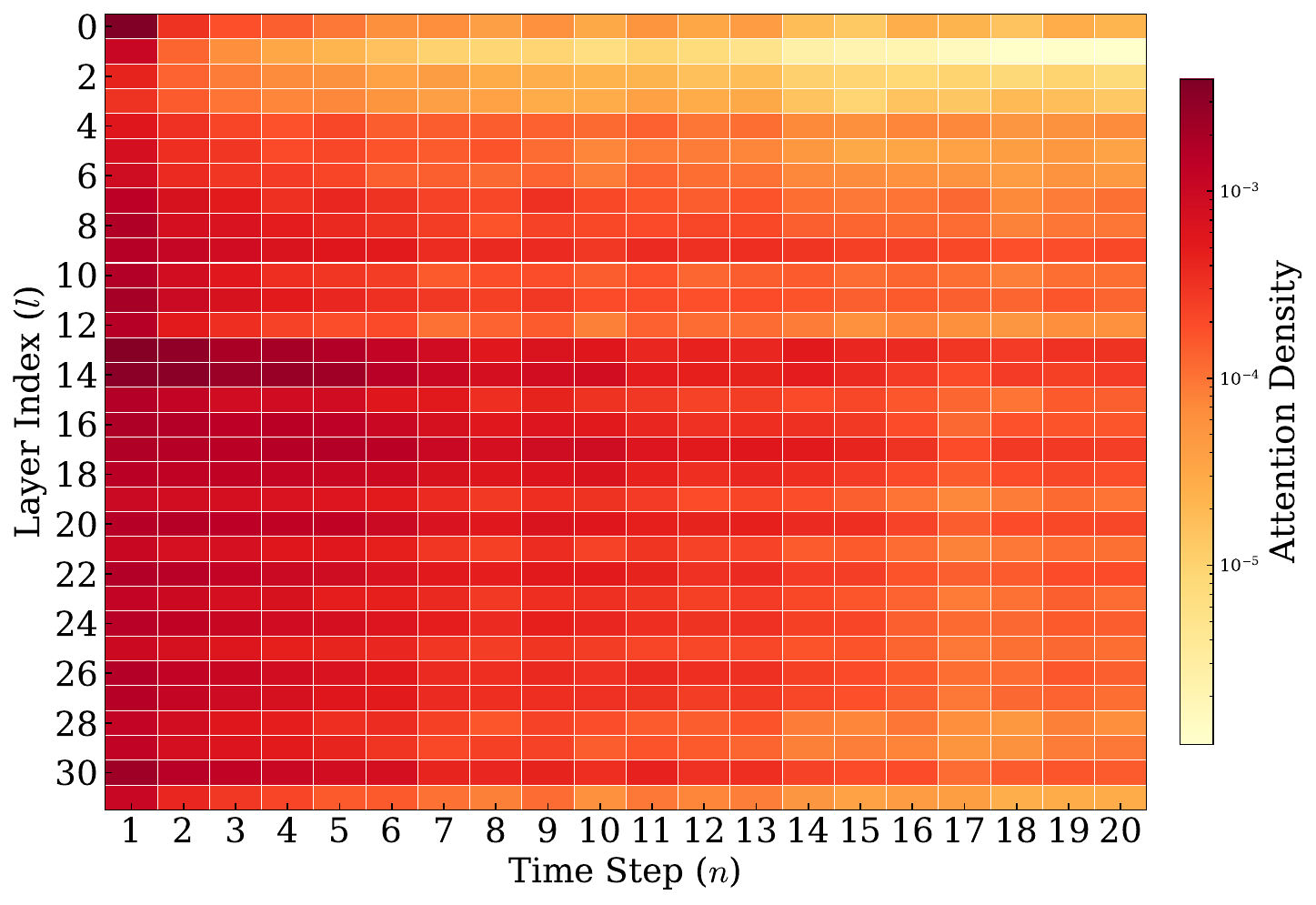}
        \caption{Attention density $\rho_1^l$ of the target knowledge fragment $x_1$ across layers $l$ and time steps $n$.}
        \label{fig:investigation_sub2}
    \end{subfigure}
    \hfill
        \begin{subfigure}[b]{0.8\linewidth}
        \centering
        \includegraphics[width=\textwidth]{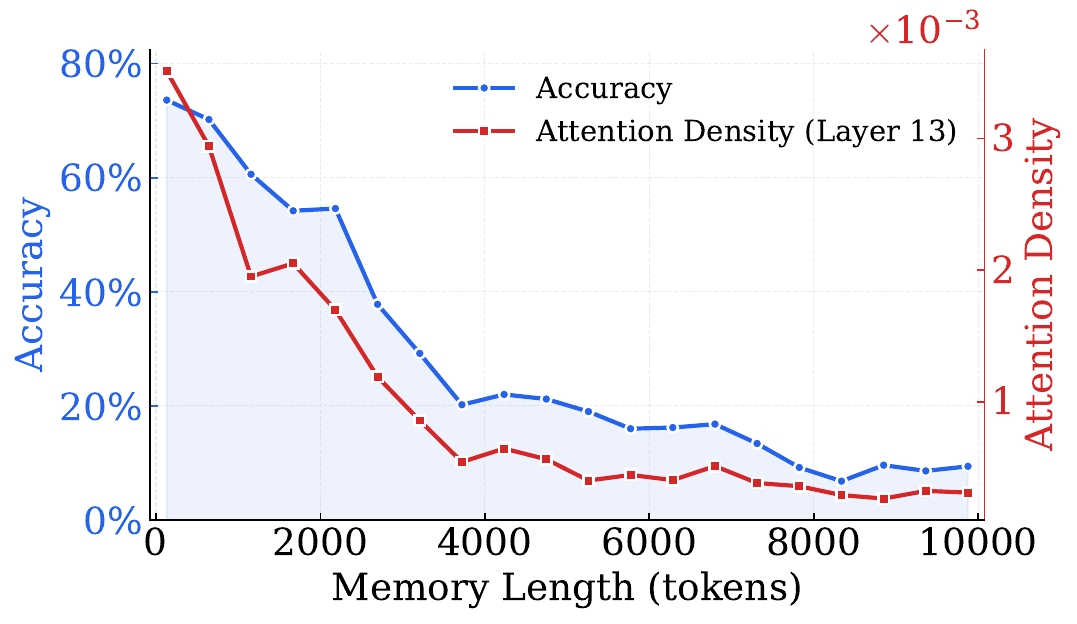}
        \caption{QA accuracy and attention density $\rho_1^{13}$ of the target memory fragment $m_1$ at layer 13.}
        \label{fig:investigation_sub4}
    \end{subfigure}
    \hfill
    \begin{subfigure}[b]{\linewidth}
        \centering
        \includegraphics[width=\textwidth]{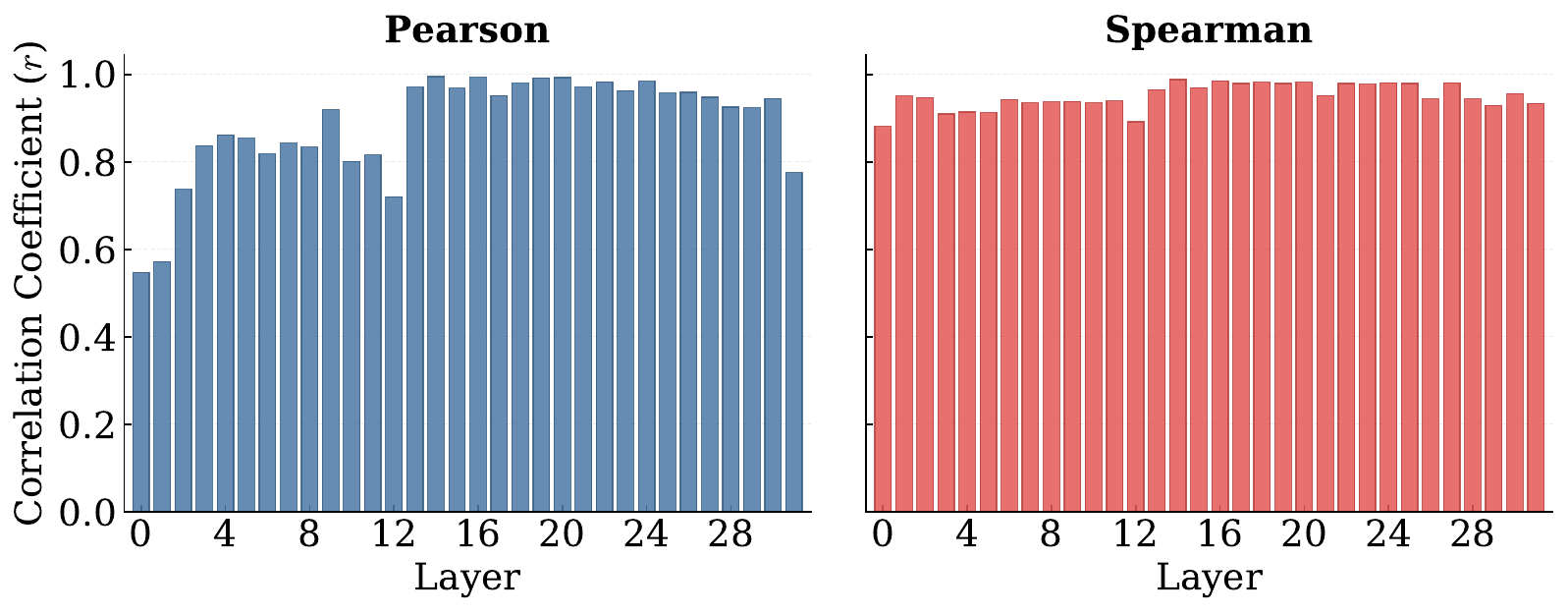}
        \caption{Per-layer Pearson and Spearman correlation coefficients between attention density $\rho_i^l$ and QA accuracy across 20 steps.}
        \label{fig:investigation_sub3}
    \end{subfigure}
    
    \caption{Effects and analyses of positional distortion and attention density in latent memory.}
    \label{fig:three_subfigures}
\end{figure}

\subsection{Effects of Positional Distortion}
\label{sec: Effects of Positional Distortion}
To investigate how positional distortion affects inference performance, we conduct experiments based on vanilla latent memory, using Llama-3.1-8B-Instruct\footnote{\url{https://huggingface.co/meta-llama/Llama-3.1-8B-Instruct}.} with greedy decoding \texttt{(max\_new\_tokens = 32)}.
The dataset used is NaturalQuestions (NaturalQA)~\cite{kwiatkowski-etal-2019-natural}\footnote{We use a filtered subset of the dataset to ensure low data leakage and mutual independence among contexts: \url{https://huggingface.co/datasets/YuWangX/KnowledgeRetentionProcessed}.}, where each item is a \texttt{(context, question, answer)} triple.
We sample 500 groups for evaluation. Within each group, the protocol (largely following the setup of MemoryLLM and M+) is defined as follows:

(1)  Each context serves as the knowledge $x_i$ ($|x_i| \leq 512$) stored at time step $i$.

(2) After every memory update that obtains $\theta_i$ ($1 \leq i \leq 20$), the first knowledge $x_1$ is evaluated.

(3) The model $\mathcal{M}_{\theta_i,\phi}$ is queried with the \texttt{question} corresponding to $x_1$. A response is deemed correct if the gold \texttt{answer} appears as a substring of the model output.

Figure~\ref{fig:investigation_sub1} shows that the accuracy drops sharply as $n$ and $N_n$ (memory length at step $n$) increase, with performance collapsing once $N_n$ exceeds 2000. It indicates that positional distortion of latent memory is much more severe than the degradation observed in the phenomenon~\cite{liu-etal-2024-lost} of long contexts where positional encodings are correctly assigned throughout the input sequence.
Experiments and analyses based on the setting \textit{without} positional distortion are shown in Appendix~\ref{app:Analyses of Attention Density without Positional Distortion}.

\subsection{Attention Density as an Indicator}
\label{sec: investigation_attention}
Although positional distortion is a fundamental issue inherent to the latent-memory paradigm, it is not directly addressable within this framework. This motivates us to seek a quantifiable indicator correlated with performance degradation. 
To this end, we introduce layer-wise \textit{attention density} $\rho_i^l$ for each knowledge item $x_i$, which captures query-aware cruciality at the granular level of fragments.
A related but distinct line of work, exemplified by SnapKV~\cite{li2024snapkv}, uses per-token attention cruciality for  KV-cache compression rather than long-term memory retention.
\subsubsection{Attention Density}
\label{sec:attention-density}
We define a per-layer, per-fragment relevance score
$\rho_i^l$ called the \emph{attention density} of the stored
knowledge fragment $x_i$ at layer $l$, which measures how much
attention the current prompt $p$ pays to $x_i$.

Let $l$ denote a designated transformer layer. At inference time, the memory is stored as a \emph{per-layer prefix}: the layer-$l$ input is formed by prepending the \texttt{bos} token and the stored memory hidden states to the prompt's layer-$l$ states,
\begin{equation}
\label{eq:concat-seq}
    H^{l} = [\texttt{bos}^{l};\, m_1^{l};\, \dots;\, m_n^{l};\, H_p^{l}]
    \in \mathbb{R}^{S \times d},
\end{equation}
with $S = 1 + N_n + |p|$ and $N_n = \sum_{i=1}^{n} L_i$ the total
number of retained memory positions (where $L_i$ is the current retained length of $x_i$). Since only the prompt tokens
need to query the memory, we run multi-head self-attention on
$H^{l}$ under a causal mask, restrict the queries to the $|p|$
prompt positions, and average across heads. This yields a
head-averaged \emph{prompt-to-sequence} attention matrix
$\bar{A}^{l} \in \mathbb{R}^{|p| \times S}$ (full derivation in
Appendix~\ref{appendix:attention-density-derivation}). Taking its
last row gives the per-position attention score from the prompt's
last token, $\bar{a}_t^{l} = \bar{A}_{|p|,\, t}^{l}$.

We then collapse these per-position scores into a single
relevance score for each fragment $x_i$ by averaging over the
positions $P_i$ that $x_i$ occupies in the concatenated memory
(excluding \texttt{bos}):
\begin{equation}
\label{eq:density}
    \rho_i^l \;=\; \frac{1}{L_i} \sum_{t \in P_i} \bar{a}_t^l,
    \qquad i = 1, \dots, n,
\end{equation} 
Intuitively, $\rho_i^l$ summarizes how strongly layer $l$ routes the prompt's attention toward the memory region occupied by $x_i$.\footnote{We also consider an all-token variant that averages $\bar{a}_t^l$ over all prompt positions rather than only the last one; details are in Appendix~\ref{appendix:Variants of Attention Computation}.}

\subsubsection{Analyses of Attention Density}
\label{sec:Analyses of Attention Density}

Figure~\ref{fig:investigation_sub2} presents a heatmap of the attention density $\rho_1^l$ for the target knowledge across each layer $l$ and time step $n$, following the setup described in Section~\ref{sec: Effects of Positional Distortion}. A clear trend emerges: in several layers (particularly layers 13–20) attention density decreases markedly as the time step increases.
Figure~\ref{fig:investigation_sub4} uses two curves to show that QA accuracy is highly aligned with the attention density allocated to the target memory fragment at layer 13.

To quantify the layer-wise correlation between accuracy degradation and attention density reduction, we compute the Pearson and Spearman correlation coefficients at each layer $l$, as shown in Figure~\ref{fig:investigation_sub3}.\footnote{The two coefficients are complementary: Pearson captures linear co-variation, while Spearman, being rank-based, is robust to the heavy-tailed distribution of attention density and reflects monotonic (but possibly non-linear) dependence.}
Notably, both coefficients exceed 0.95 across the majority of layers 13-20, indicating a near-perfect monotonic relationship between attention density reduction and accuracy decline.
This strong correlation suggests that attention density is a key indicator associated with accuracy.

Moreover, according to Eq.~\ref{eq:density} and its previous derivation, a straightforward approach to increasing the attention density (correlated with the QA performance) of the target memory fragment is to remove non-target ones (temporarily) during inference.
In this case, \textit{the first step is to trace and identify the target memory fragment}.

\begin{figure*}[!t]
    \centering
    
    \begin{subfigure}[b]{0.25\textwidth}
        \centering
        \includegraphics[width=\textwidth]{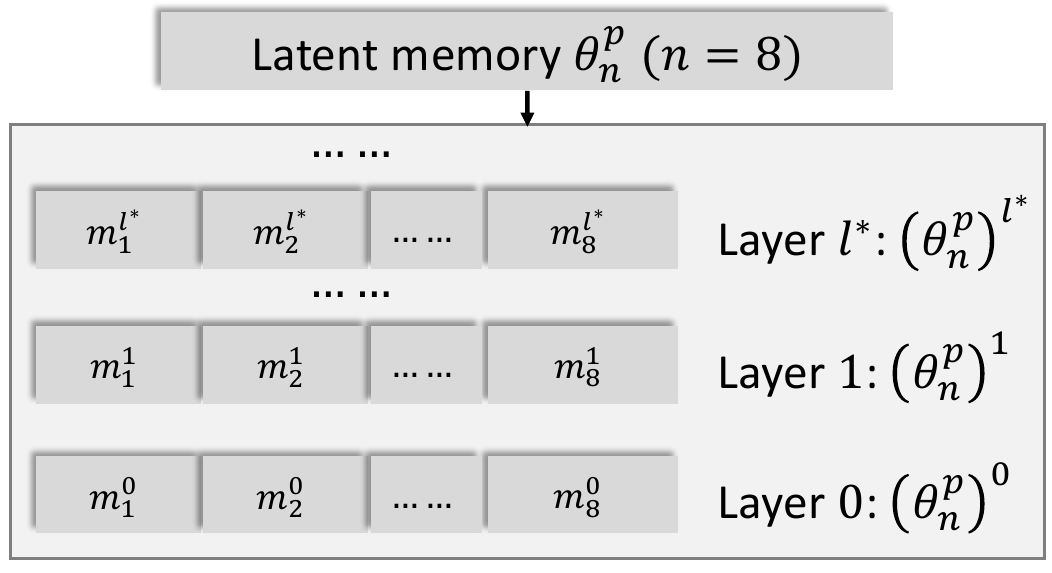}
        \caption{The stored latent memory before defragmentation.}
        \label{fig:MemDefrag_sub1}
    \end{subfigure}
    \hfill
    \begin{subfigure}[b]{0.45\textwidth}
        \centering
        \includegraphics[width=\textwidth]{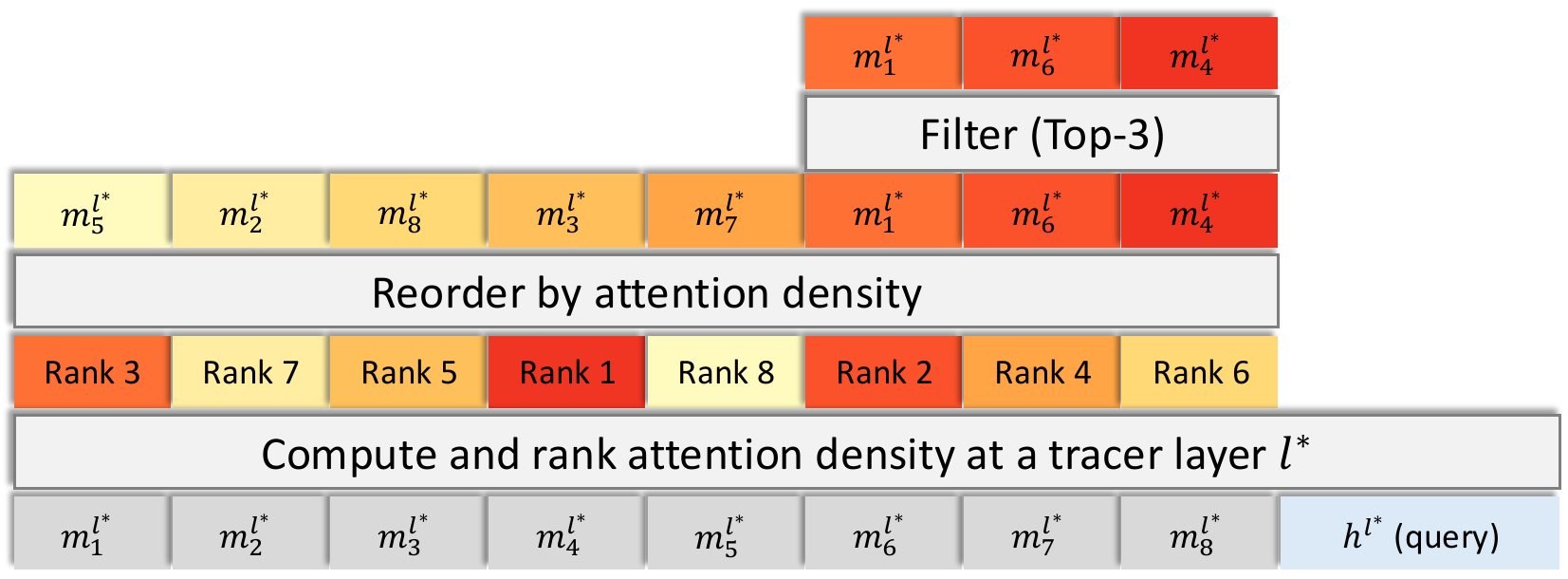}
        \caption{Defragmentation, which involves ranking, reordering, and filtering memory fragments).}
        \label{fig:MemDefrag_sub2}
    \end{subfigure}
    \hfill
    \begin{subfigure}[b]{0.25\textwidth}
        \centering
        \includegraphics[width=\textwidth]{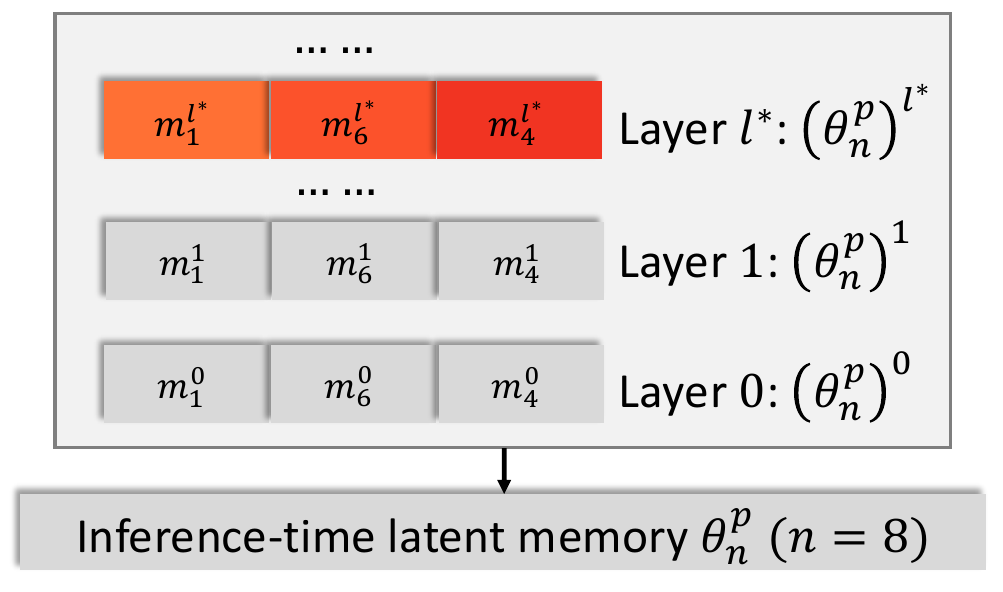}
        \caption{The defragmented latent memory for inference.}
        \label{fig:MemDefrag_sub3}
    \end{subfigure}
    
    \caption{An example for the defragmentation processes of our MemDefrag framework, in which the time step is 8 and Top-3 filtering strategy is adopted. At each layer, the defragmented latent memory follows the ordered and filtered fragment indexes at the tracer layer $l^*$.}
    \label{fig:MemDefrag}
\end{figure*}

\section{Attention Density as a Tracing Signal}
\label{sec: Attention Density as Retrieval Signal}
In this section, we provide an insight that attention density serves not only as an indicator correlated with performance, but also as a \textbf{tracing signal}: \textit{Certain layers exhibit an inherent tracing capability, consistently identifying the target fragment through attention density ranking}.
Our insight can be illustrated by the following experiments.

We evaluate 500 groups from the NaturalQA dataset. For each group, 20 knowledge fragments are stored, where the target fragment is placed at each of the 20 positions in turn (from first to last) to mitigate positional bias. Table~\ref{tab:retrieval_layers} reports the average attention density rank of the target fragment across all positions for each layer. Layer 13 achieves the lowest mean rank of 1.66, with a Top-1 accuracy of 85.6\% and a Top-3 accuracy of 95.7\%, indicating that it consistently assigns the highest attention density to the target fragment regardless of its position. 

Notably, this aligns with findings in activation steering studies~\cite{turner2024steeringlanguagemodelsactivation, panickssery2024steeringllama2contrastive, zou2025representationengineeringtopdownapproach}, which identify specific middle layers as the most effective intervention point to control model behaviors.
One possible explanation is that intermediate layers can encode richer representations than both earlier layers and final layers, holding implications for feature relevance and extraction~\cite{pmlr-v267-skean25a, li2026repolanguagemodelscontext}.
Further, analyses about the tracing capabilities based on the setting \textit{without} positional distortion are shown in Appendix~\ref{app:Analyses of Tracing Capacities without Positional Distortion}.


\begin{table}[!t]
\centering
\resizebox{\linewidth}{!}{
\begin{tabular}{c|c|ccccc}
\toprule
Layer & $\overline{\text{Rank}}\downarrow$ & Top-$1\uparrow$ & Top-$2\uparrow$ & Top-$3\uparrow$ & Top-$4\uparrow$ & Top-$5\uparrow$ \\
\midrule
13 & 1.66 & 85.6 & 94.7 & 95.7 & 96.3 & 96.6 \\
9  & 2.14 & 78.3 & 84.9 & 92.2 & 93.2 & 93.8 \\
14 & 3.19 & 75.1 & 83.2 & 85.3 & 86.2 & 86.9 \\
17 & 4.34 & 68.2 & 76.7 & 78.1 & 79.2 & 79.9 \\
11 & 5.01 & 60.7 & 72.3 & 73.9 & 74.7 & 75.5 \\
\bottomrule
\end{tabular}}
\caption{Layers with the highest tracing capability based on 32-layer Llama-3.1-8B-Instruct, NaturalQA and last-token attention, ranked by average attention density rank, with Top-$K$ accuracies (\%) across all positions.}
\label{tab:retrieval_layers}
\end{table}

\section{MemDefrag}
We propose \textbf{MemDefrag} of which latent memory defragmentation, given a prompt, traces and consolidates the most relevant memory fragments for inference while discarding irrelevant ones to mitigate accumulated positional distortion and to improve the attention density allocated to the relevant fragments.
Tracing the relevant fragment(s) is guided by attention density at one \textit{tracer layer} $l^*$ (such as layer 13 or 9 listed in Table~\ref{tab:retrieval_layers} in Section~\ref{sec: Attention Density as Retrieval Signal}).

As MemDefrag is generally applicable to latent memory designs, \textit{without loss of generality}, we present it as an extension of the memory formation and concatenation-based memory evolution in vanilla latent memory (Section~\ref{sec: Vanilla Latent Memory}).

\subsection{Inference-Time Memory Defragmentation}
\label{sec:Inference-Time Memory Defragmentation}
At inference time, given a prompt $p$ and the accumulated memory $\theta_n = \{m_1, \dots, m_n\}$, the function $f_{\text{defrag}}(\theta_n, p)$ defragments them to a prompt-specific inference-time temporary memory $\theta_n^p$. 

To defragment the whole memory $\theta_n$, all $n$ memory fragments are first sorted by their attention density in ascending order:
\begin{equation}
    \pi = \mathrm{argsort}(\rho^{l^*}_1, \dots, \rho^{l^*}_n),
\end{equation}
such that $\rho^{l^*}_{\pi(1)} \leq \rho^{l^*}_{\pi(2)} \leq \cdots \leq \rho^{l^*}_{\pi(n)}$. The entire memory $\theta_{n}$ is then reordered by $\pi$ at \textit{each} layer:
\begin{equation}
    \tilde{\theta}_n^l = [m_{\pi(1)}^l;\, m_{\pi(2)}^l;\, \dots;\, m_{\pi(n)}^l], \ l = 1, \dots, L.
\end{equation}
This places the most relevant memory fragments at the end (i.e., closest to the prompt), thereby leveraging the recency bias of causal attention.

Finally, we filter only the Top-$K$ most relevant memory fragments by truncating the reordered sequence from the front:
\begin{equation}
    \theta_n^p = \{m_{\pi(n-K+1)},\, \dots,\, m_{\pi(n)}\},
\end{equation}
i.e., the hidden states at each layer are:
\begin{equation}
    (\theta_n^p)^l = [m_{\pi(n-K+1)}^l;\, \dots;\, m_{\pi(n)}^l] \in \mathbb{R}^{N_K \times d},
\end{equation}
where $N_K = \sum_{k=n-K+1}^{n} L_{\pi(k)}$, in which $L_i$ denotes the current retained length of $x_i$. The prompt-specific memory $\theta_n^p$ is then used for inference: $\text{output} = \mathcal{M}_{\theta_n^p, \phi}(p)$.
Algorithm~\ref{alg:memory_retrieval} shows the process of defragmenting $\theta_n^p$ from $\theta_n$.

\begin{algorithm}[t]
\small
\caption{Memory Defragmentation}
\label{alg:memory_retrieval}
\begin{algorithmic}[1]
\Input Accumulated memory $\theta_n = \{m_1, \dots, m_n\}$ with attention density $\{\rho^{l^*}_1, \dots, \rho^{l^*}_n\}$ of a tracer layer $l^*$ based on a prompt $p$, number of retained knowledge $K$
\Output Inference-time temporary memory $\theta_n^p$

\State $\pi \gets \mathrm{argsort}(\rho^{l^*}_1, \dots, \rho^{l^*}_n)$ \Comment{Ascending order}
\For{each layer $l = 1, \dots, L$}
    \State $\tilde{\theta}_n^l \gets [m_{\pi(1)}^l;\, m_{\pi(2)}^l;\, \dots;\, m_{\pi(n)}^l]$
\EndFor

\State $\theta_n^p \gets \{m_{\pi(n-K+1)}, \dots, m_{\pi(n)}\}$ \Comment{Filter Top-$K$ relevant memories}

\State \Return $\theta_n^p$
\end{algorithmic}
\end{algorithm}

\subsection{Memory Evolution with Informativeness-Guided Forgetting}
Beyond defragmentation,  our MemDefrag further incorporates an informativeness-based forgetting mechanism for memory evolution (e.g., updating $\theta_{n-1}$ to $\theta_n$).
The presence of forgetting is essential for latent memory, as  practical deployment constraints demand an effective eviction strategy.

Unlike MemoryLLM~\cite{10.5555/3692070.3694135} and M+~\cite{wang2025m} that randomly drop memory states independently at each layer, indiscriminately losing information, MemDefrag introduces a selective strategy that operates coherently across all layers of each token.

\subsubsection{Sketch of Memory Evolution}
Given the current length $N_{n-1}$, the formed memory $m_{n}$ of new knowledge $x_{n}$, and a maximum length $N_{\text{max}}$ (set empirically), $N_{\text{forget}} \ (N_{\text{forget}} = N_{n-1} + |x_{n}| - N_{\text{max}})$ hidden states are needed to be dropped. 
The evolution proceeds in two cases: $\theta_n = U_{\phi}(\theta_{n-1}, f_{\text{form}}(x_n)) = U_{\phi}(\theta_{n-1}, m_n)$ equals $\text{concat}(\theta_{n-1}, m_n)$ when $N_{\text{forget}} \leq 0$, and equals $\text{concat}(f_{\text{forget}}(\theta_{n-1}, N_{\text{forget}}), m_n)$ when $N_{\text{forget}} > 0$, where $|x_n| \leq N_{\text{max}}$, and the function $f_{\text{forget}}(\cdot)$ prunes $N_{\text{forget}}$ states from $\theta_{n-1}$ to store the incoming memory.
Algorithm~\ref{alg:memory_evolution} in Appendix sketches the memory evolution with the forgetting mechanism.

\subsubsection{Informativeness-Based Proportional Forgetting of Memory}
\label{sec:Informativeness-Based Proportional Forgetting of Memory}
Our forgetting function $f_\text{forget}(\cdot)$ performs \textit{proportional forgetting}: when $N_\text{forget}$ hidden states need to be dropped, the quota is distributed proportionally across all stored memory fragments. 

Inspired by the ideas proposed in \textit{Selective Context}~\cite{li-etal-2023-compressing} and other techniques of prompt compression~\cite{jiang-etal-2023-llmlingua, pan-etal-2024-llmlingua}, which identify and prune redundant tokens, $f_\text{forget}(\cdot)$ prunes hidden states within each stored memory fragment based on the informativeness of the corresponding tokens. Detailed descriptions of this forgetting strategy is shown in Appendix~\ref{app:Descriptions of Informativeness-Based Proportional Forgetting}.

\begin{figure}[!t]
    \centering
    \includegraphics[width=\linewidth]{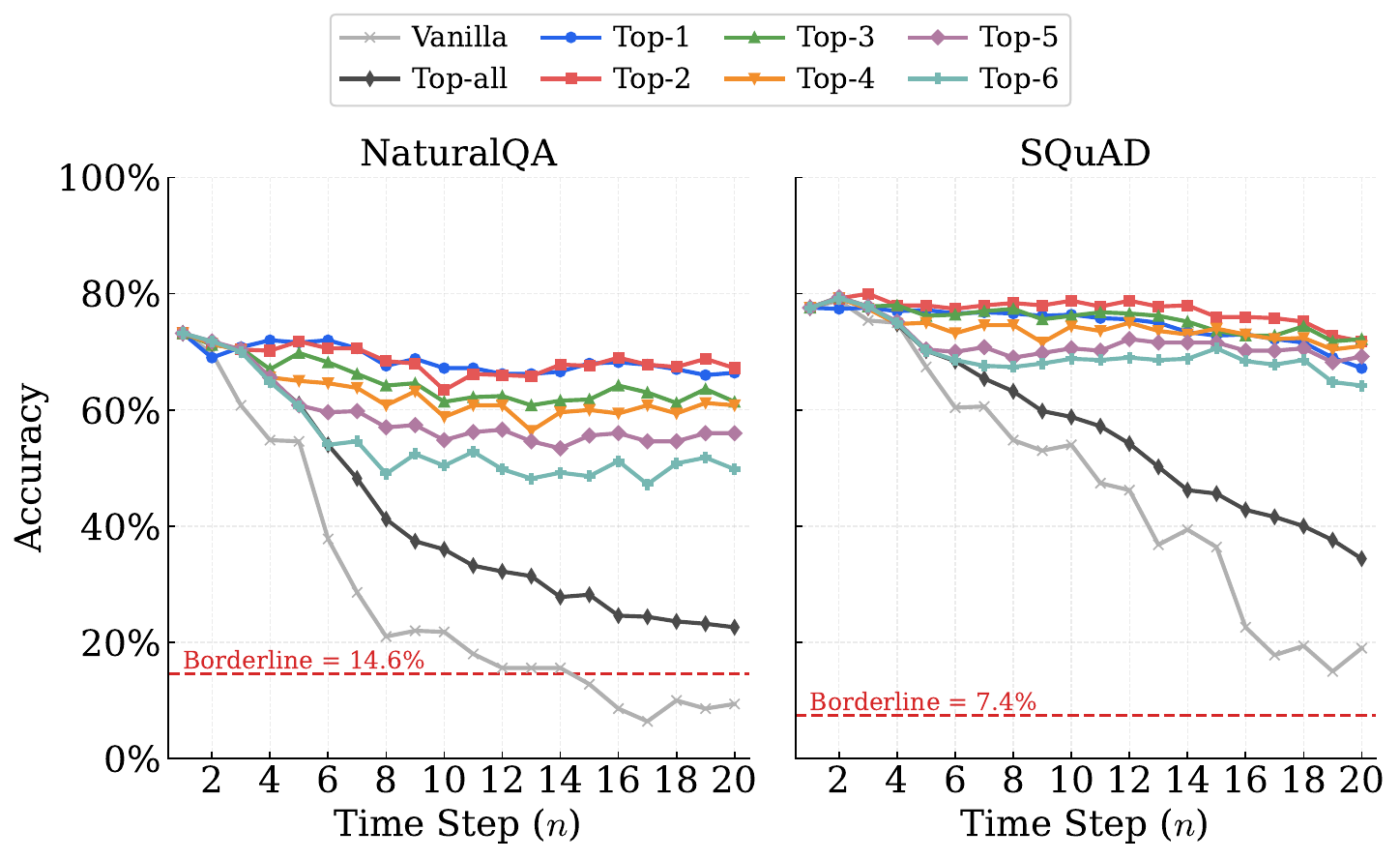}
    \caption{20-step knowledge retention on NaturalQA and SQuAD. \textit{Vanilla} refers to the vanilla concatenation-based latent memory introduced in Section~\ref{sec: Vanilla Latent Memory}. All other variants are instances of memory defragmentation (Section~\ref{sec:Inference-Time Memory Defragmentation}): in which \textit{Top-all} applies reordering alone without truncation, while Top-$K$ filters only the $K$ highest-ranked memory fragments after reordering.}
    \label{fig:retention_different_k}
\end{figure}

\section{Experiments}
\label{sec: Experiments}
We adopt MemoryLLM~\cite{10.5555/3692070.3694135} and M+~\cite{wang2025m}, latent-memory methods that explicitly account for forgetting and knowledge retention, as baselines for our MemDefrag. Following their evaluation protocols, we assess MemDefrag across knowledge retention (Section~\ref{sec:Knowledge Retention Experiments}) and long-context QA benchmarks (Section~\ref{sec:Long Context Experiments}). 
We also provide analyses on generality and compatibility on various aspects (Section~\ref{sec:Generality across Models}). 

Memoryllm-8b\footnote{\url{https://huggingface.co/YuWangX/memoryllm-8b}} and M+-8b\footnote{\url{https://huggingface.co/YuWangX/mplus-8b}} serve as baseline models. Although MemDefrag is model-agnostic and applicable to backbone architectures, we conduct all comparative experiments (in Section~\ref{sec:Knowledge Retention Experiments} and~\ref{sec:Long Context Experiments}) on Llama-3.1-8B-Instruct for consistency. All three models are built upon Llama-3.1-8B~\citep{grattafiori2024llama3herdmodels}, ensuring a fair comparison.
The maximum length $N_{\text{max}}$ of MemDefrag is set as 12,800, which is equal to the memory token numbers of Memoryllm-8b and M+-8b.
The layer 13 serves as the tracer layer of MemDefrag because of its highest tracing capability in Llama-3.1-8B-Instruct discussed in Section~\ref{sec: Attention Density as Retrieval Signal} and Appendix~\ref{appendix:Variants of Attention Computation}.
All experiments are conducted on $4 \times$ GPUs that feature 141 GB of HBM3e memory.

\begin{figure}[!t]
    \centering
    \includegraphics[width=\linewidth]{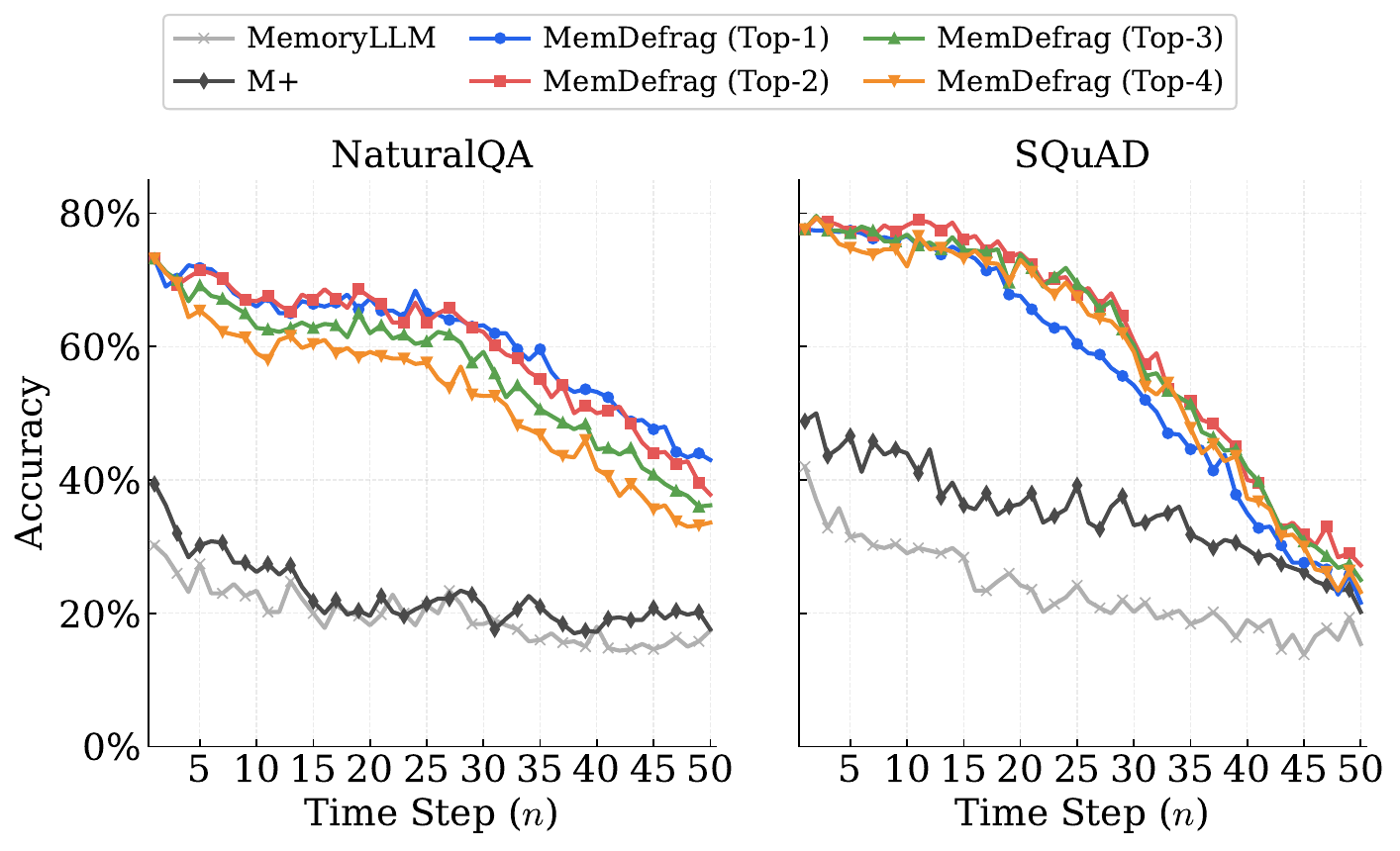}
    \caption{50-step knowledge retention on NaturalQA and SQuAD, comparing MemDefrag (Top-1 through Top-4 with informativeness-based forgetting) against MemoryLLM and M+.}
    \label{fig:retention_comparison}
\end{figure}

\subsection{Knowledge Retention Experiments}
\label{sec:Knowledge Retention Experiments}
To evaluate the ability of MemDefrag to recall long-term knowledge, we follow the setup in MemoryLLM and M+ on datasets NaturalQA~\cite{kwiatkowski-etal-2019-natural} and SQuAD~\cite{rajpurkar-etal-2016-squad}, where each item is a \texttt{(context, question, answer)} triple.
The implementation details follow Section~\ref{sec: Effects of Positional Distortion}.
We calculate last-token attention for NaturalQA and all-token attention for SQuAD as discussed in Appendix~\ref{appendix:Variants of Attention Computation}.
To eliminate positional bias of evaluation, i.e., to prevent the target fragment from consistently occupying the first position, the stored memory fragments are shuffled after each memory update.
For each dataset, 500 groups are sampled for evaluation.\footnote{Note that the sampled groups used here are disjoint from those in Section~\ref{sec:Investigation}, ensuring that hyperparameter selection is not biased by prior observations on the same data.}
Besides, we report the QA accuracy obtained without storing any memory fragments (NaturalQA: 14.6\%; SQuAD: 7.4\%), referred to as the \textit{borderline} accuracy.

\subsubsection{Ablation Study of Different \texorpdfstring{$K$}{K}}
\label{sec: Ablation Study of Different K}

We first study the effects of different $K$, which is used to keep Top-$K$ memory fragments for inference-time memory Defragmentation (Section~\ref{sec:Inference-Time Memory Defragmentation}). The maximum time step is 20. As the maximum memory length $N_{20}$ is less than $N_{\text{max}} = 12,800$, which does not trigger any forgetting strategy. Figure~\ref{fig:retention_different_k} shows the results of knowledge retention, which reveals several key findings:

(1) Vanilla latent memory suffers from severe degradation as the number of time steps increases: accuracy drops from 73.2\% to below 9.4\% on NaturalQA and from 77.6\% to 19.0\% on SQuAD. 
On NaturalQA, vanilla accuracy eventually falls below the borderline, indicating that the accumulated positional distortion can compromise even the standard inference capacity of the language model.

(2) All MemDefrag variants substantially slow down the degradation curve compared to the vanilla. Even Top-all, which only reorders fragments without filtering, maintains higher accuracy throughout the 20 steps (e.g., 22.6\% vs.\ 9.4\% at step 20 on NaturalQA), demonstrating that solely placing the target memory fragment at a position closer to the prompt contributes to performance improvement.

(3) Top-2 achieves the highest accuracy on NaturalQA (67.2\%) at step 20, while Top-3 achieves the highest on SQuAD (72.2\%) at step 20. 
Top-6 (as well as Top-all) achieves lower accuracy in both datasets, revealing a trade-off: \textit{a larger $K$ improves the chance of tracing the target memory fragment but reduces its attention density}. Attention density is highly correlated with QA accuracy (discussed in Section~\ref{sec:Analyses of Attention Density} and Figure~\ref{fig:investigation_sub3}).

\subsubsection{Ablation Study of Forgetting Strategies}
We study the differences of forgetting strategies: \textit{informativeness-based} (Section~\ref{sec:Informativeness-Based Proportional Forgetting of Memory}) vs. \textit{random}  proportional forgetting.
The detailed results to validate the informativeness-based forgetting are shown in Appendix~\ref{app:Supplementary Results of Ablation Study of Forgetting Strategies}.

\begin{table}[t]
\centering
\resizebox{\linewidth}{!}{%
\begin{tabular}{l ccccc ccccc}
\toprule
& \multicolumn{5}{c}{\textbf{NaturalQA}} & \multicolumn{5}{c}{\textbf{SQuAD}} \\
\cmidrule(lr){2-6} \cmidrule(lr){7-11}
\textbf{Method} & $n$=10 & $n$=20 & $n$=30 & $n$=40 & $n$=50 & $n$=10 & $n$=20 & $n$=30 & $n$=40 & $n$=50 \\
\hline
\rowcolor{blue!8} \multicolumn{11}{c}{\textit{Baselines}} \\ \hline
MemoryLLM & 23.4 & 18.2 & 18.4 & 18.0 & 17.4 & 29.0 & 24.2 & 20.2 & 19.0 & 15.4 \\
M+        & 26.2 & 19.6 & 21.0 & 17.2 & 17.6 & 44.0 & 36.4 & 33.2 & 29.6 & 20.2 \\
\hline
\rowcolor{blue!8} \multicolumn{11}{c}{\textit{MemDefrag}} \\ \hline
Top-1     & 66.0 & 67.2 & \textbf{63.2} & \textbf{53.2} & \textbf{43.0} & 76.6 & 67.6 & 54.2 & 35.0 & 21.6 \\
Top-2     & \textbf{66.8} & \textbf{67.4} & 62.2 & 50.0 & 37.8 & \textbf{78.2} & \textbf{74.0} & \textbf{60.8} & 40.0 & \textbf{27.2} \\
Top-3     & 62.8 & 62.0 & 59.2 & 44.6 & 36.2 & 76.8 & 73.8 & 60.2 & \textbf{41.6} & 25.0 \\
Top-4     & 59.0 & 59.2 & 52.6 & 41.6 & 33.6 & 72.0 & 73.0 & 59.2 & 37.2 & 23.2 \\
\bottomrule
\end{tabular}%
}
\caption{QA accuracy (\%) of knowledge retention at selected time steps on NaturalQA and SQuAD. 
\textbf{Bold} indicates the best result in each column.}
\label{tab:retention_comparison}
\end{table}

\begin{table}[!t]
\centering
\resizebox{\linewidth}{!}{%
\begin{tabular}{lcccc}
\toprule
 \textbf{Method} & \textbf{String match} & \textbf{F1} & \textbf{ROUGE-L} & \textbf{BERTScore} \\
\hline
\rowcolor{blue!8} \multicolumn{5}{c}{\textit{2WikiMultihopQA}} \\ \hline
MemoryLLM            & 35.00 & 9.21 & \textbf{10.13} & 82.32 \\
M+                   & 35.60 & 9.41 & 9.79           & 81.81 \\
MemDefrag (Top-2)    & \textbf{36.00} & \textbf{9.51} & 9.53  & \textbf{83.82} \\
\hline
\rowcolor{blue!8} \multicolumn{5}{c}{\textit{HotpotQA}} \\ \hline
MemoryLLM            & 37.50 & 8.10 & 8.23  & 83.15 \\
M+                   & 38.00 & 8.56 & 8.34 & 82.84 \\
MemDefrag (Top-2)    & \textbf{39.00} & \textbf{8.88} & \textbf{8.43} & \textbf{84.22} \\
\hline
\rowcolor{blue!8} \multicolumn{5}{c}{\textit{MuSiQue}} \\ \hline
MemoryLLM            & 16.00 & 5.08 & 5.16          & 81.38 \\
M+                   & 19.50 & 6.03 & \textbf{5.84} & 81.37 \\
MemDefrag (Top-2)    & \textbf{20.00} & \textbf{6.16} & 5.65 & \textbf{83.49} \\
\hline
\rowcolor{blue!8} \multicolumn{5}{c}{\textit{MultiFieldQA-en}} \\ \hline
MemoryLLM            & 3.33           & 14.76 & 14.83 & 85.92 \\
M+                   & \textbf{6.00}  & 14.11 & 14.41 & 85.06 \\
MemDefrag (Top-2)    & \textbf{6.00}  & \textbf{15.74} & \textbf{15.51} & \textbf{85.37} \\
\hline
\rowcolor{blue!8} \multicolumn{5}{c}{\textit{NarrativeQA}} \\ \hline
MemoryLLM            & 9.84  & 3.37 & 3.79 & 81.27 \\
M+                   & 4.92  & 3.07 & 3.63 & 81.60 \\
MemDefrag (Top-2)    & \textbf{11.48} & \textbf{7.81} & \textbf{7.77} & \textbf{82.69} \\
\hline
\rowcolor{blue!8} \multicolumn{5}{c}{\textit{Qasper}} \\ \hline
MemoryLLM            & 17.77 & 8.58 & 8.14 & 81.81 \\
M+                   & 17.26 & 8.34 & 8.34 & 81.36 \\
MemDefrag (Top-2)    & \textbf{18.27} & \textbf{9.29} & \textbf{8.66} & \textbf{82.36} \\
\bottomrule
\end{tabular}%
}
\caption{LongBench results ($\times 100$) comparison across three models. Best results per dataset are in \textbf{bold}.}
\label{tab:longbench}
\end{table}

\subsubsection{Comparison with MemoryLLM and M+}
In MemoryLLM and M+, each 512-token chunk is compressed into 256 memory vectors per layer in a fixed 2:1 ratio.\footnote{To ensure a fair comparison that accounts for this compression, each knowledge fragment $x_i$ injected into MemoryLLM and M+ is 512 tokens.} 
Figure~\ref{fig:retention_comparison} presents the results comparing MemoryLLM, M+ and, our MemDefrag from Top-1 to Top-4 memory defragmentation with informativeness-based forgetting. Several findings stand out:

(1) All MemDefrag variants substantially and consistently outperform both MemoryLLM and M+ across the 50 steps on both datasets. Table~\ref{tab:retention_comparison} summarizes the accuracy at $\{10,20,30,40,50\}$ time steps. 
By step 50, Top-1 retains an accuracy of 43.0\% on NaturalQA vs. 17.4\% for MemoryLLM and 17.6\% for M+. 

(2) The optimum $K$ remains small and does not increase with the number of steps. This suggests that concentrating attention density on the traced target fragment (achieved by filtering fewer candidates) can be more beneficial than broadening the candidate pool to increase the likelihood of including the target.

(3) Even at the earliest time steps, MemDefrag substantially outperforms MemoryLLM and M+, despite all three methods being built upon the same Llama-3.1-8B backbone.  
This indicates that: The fixed compression of MemoryLLM and M+ incurs a non-trivial information loss during memory formation, leading to a significant accuracy reduction before any long-term degradation occurs.

Further, Appendix~\ref{app:Efficiency Analyses} analyzes the running-time efficiency of conducting an experimental group which includes 50 memory fragment formation along with per-step evaluation.

\subsection{Long Context Experiments}
\label{sec:Long Context Experiments}
In this section, we evaluate the long-context modeling capabilities of MemDefrag. Following the experiment settings of MemoryLLM and M+, six datasets of the \textit{LongBench}~\cite{bai-etal-2024-longbench} are adopted, where the maximum length of each item is 16,384 tokens.
Four metrics are used for evaluation: (1) String match following the rule in Section~\ref{sec: Effects of Positional Distortion}, (2) F1 score (3) ROUGE-L~\cite{lin-2004-rouge}, and (4) BERTScore~\cite{Zhang*2020BERTScore:}.
Detailed settings are shown in Appendix~\ref{app:Setups for Long Context Experiments}. 

Table~\ref{tab:longbench} reports the results, of which Top-2 memory defragmentation for MemDefrag is adopted, and the attention computation is based on all prompt tokens (Appendix~\ref{appendix:Variants of Attention Computation}). MemDefrag consistently outperforms MemoryLLM and M+ in the majority of datasets and metrics, achieving the best results in 20 out of 24 dataset-metric combinations. 
It suggests that MemDefrag effectively preserves long-context information and generalizes well across diverse question-answering scenarios.

\subsection{Generality and Compatibility}
\label{sec:Generality across Models}
\paragraph{Generality across Models.}
Beyond Llama-3.1-8B-Instruct, we study the effects of MemDefrag in another three language models: Qwen2.5-7B-Instruct~\cite{qwen2.5}, Mistral-7B-Instruct-v0.3~\cite{jiang2023mistral7b}, Gemma-2-9b-it~\cite{gemma_2024} (see Appendix~\ref{app:Knowledge Retention Experiments Based on Various Models}). 

\paragraph{Compatibility with Latent-Memory Variants.} We study the compatibility of the inference-time memory defragmentation (Section~\ref{sec:Inference-Time Memory Defragmentation}) with MemoryLLM and M+ (see Appendix~\ref{app:Compatibility with Latent-Memory Variants}).

\paragraph{Compatibility with Prompt Compression.} MemDefrag supports flexible-length memory fragments, making it compatible with prompt compression. We study the effects of compressing knowledge fragments (see Appendix~\ref{app:Compatibility with Prompt Compression}).



\section{Conclusion}
We identified two long-standing obstacles that cause degradation under repeated updates in the long-term latent memory paradigm: positional-encoding misalignment and the absence of a tracing mechanism. Through a layer-wise probing of attention density, we showed that a small set of middle transformer layers consistently concentrates the highest density on the target fragment, exposing a reliable tracing signal. Building on this signal, we proposed MemDefrag, which (1) defragments the stored memory by ranking, reordering, and filtering fragments at inference time, and (2) operates an informativeness-guided proportional forgetting strategy.
The experiments show the superiority of MemDefrag compared to MemoryLLM and M+. As the tracing signal and the defragmentation procedure are decoupled from any specific backbone, they are transferable to other models and latent-memory variants. More broadly, according to the analyses in Appendix~\ref{app:Analyses of Tracing Capacities without Positional Distortion}, \textit{defragmentation based on attention density is also potentially applicable to plain long-context inference}.

\section*{Limitations}
Top-$K$ filtering is chosen empirically and held static throughout inference, rather than being predicted per prompt and per step. A natural remedy would be to make $K$ adaptive, for instance by selecting it from the shape of the per-fragment attention-density distribution at the tracer layer. In practice, however, we find that the benefit for such an adaptive scheme is narrow. Across the LLMs and the QA benchmarks we study, the best accuracy is almost always concentrated at Top-1 or Top-2, which leaves very little room for an adaptive $K$ to improve on a well-chosen static K.
The one exception is Gemma-2-9b, where Top-1 deteriorates beyond roughly 20 update steps while Top-2 remains stable (Figure~\ref{fig:retention_gemma}). This suggests that an adaptive $K$ is primarily useful for coping with long-term drift on particular (model, dataset) pairs, rather than as a universal replacement for the static choice.
In this paper, therefore, we focus on a fixed small $K$, treating adaptive $K$ as an orthogonal design choice rather than a core component of MemDefrag.

\section*{Acknowledgments}

We express our gratitude to the anonymous reviewers for their valuable and insightful comments. This work was supported by LIGHTSPEED and JSPS KAKENHI Grant Number JP26KJ1382.

\bibliography{custom}

\appendix

\section{Related Work}
LLM memory serves as a cornerstone for transforming static LLMs into adaptive tools, particularly when their parameters cannot be promptly updated in response to interactive environment changes~\cite{hu2026memoryageaiagents, yu2026latentspacefoundationevolution, zhong2026semanticchunkingentropynatural}. 
Moreover, LLM memory is widely recognized as one of the most important capabilities (alongside reasoning, planning, perception, and tool use) that supports LLM-based agents~\cite{10.1145/3748302, qu2025tool, li2026timelabelcontinuousphase}. 

\subsection{LLM Memory Forms}
We categorize LLM memory into three types: (1) token-level memory, (2) parametric memory, and (3) latent memory.

Token-level memory is organized as explicit and discrete text tokens that can be persistently stored in a structured form. Substantial methods in this category typically focus on storing, summarizing, managing, and preventing the forgetting of dialog contents. 
For example, MemGPT~\cite{packer2024memgptllmsoperatingsystems} proposes an operating-system metaphor with hierarchical management, and Mem0~\cite{chhikara2025mem0buildingproductionreadyai} establishes standardized operations for storing and retrieving information from ongoing conversations. 
Although token-level memory offers great modularity and interpretability, these approaches suffer from the \textit{lost-in-the-middle} phenomenon~\cite{liu-etal-2024-lost} and context window bottleneck~\cite{wang2025m}.

Parametric memory stores information directly within the model's parameters. One major type is memory encoded in the model's original parameters.
For example, LMLM~\cite{zhao2025pretraininglimitedmemorylanguage} embeds retrievable knowledge during the pre-training phase, while CharacterGLM~\cite{zhou-etal-2024-characterglm} fine-tunes the LLM to exhibit distinct character traits.
The other type resides in auxiliary parameter sets, such as adapters and LoRA modules~\cite{hu2022lora}, which are attached alongside the base model. The primary advantage of parametric memory is that it incurs no additional inference-time overhead. However, it suffers from costly retraining requirements and the risk of catastrophic forgetting~\cite{doi:10.1073/pnas.1611835114}.

Latent memory refers to information carried implicitly in the model's internal representations, such as key-value caches, activations, and hidden states.
Most existing methods in this category operate as short-term (working) memory, primarily reusing the key-value (KV) cache as a form of latent storage~\cite{wu2022memorizing, gershman2025key}.
The most closely related to MemDefrag proposed in this paper is \emph{long-term} latent memory, as exemplified by MemoryLLM~\cite{10.5555/3692070.3694135}, M+~\cite{wang2025m}, and Memory\textasciicircum3~\cite{Yang_2024}.
They avoid repeatedly processing the full context and introduce substantially less inference latency. However, transferring information into latent representations may incur information loss or semantic bias, and training a dedicated module to generate such representations introduces additional computational overhead.

\subsection{Long-Term Latent Memory}
Our proposed MemDefrag falls into the category of long-term latent memory, which generally supports persistent knowledge retention, efficient retrieval across extended interactions, and graceful integration of new information without catastrophic forgetting. 
Our method stores information in per-layer hidden states, especially aligning with the paradigm of MemoryLLM~\cite{10.5555/3692070.3694135}, M+~\cite{wang2025m}, and NextMem~\cite{zhang2026nextmemlatentfactualmemory}.

MemoryLLM incorporates a substantial portion of self-updatable prefix hidden states (referred to as a \textit{memory pool}) enabling the model to integrate new knowledge both effectively and efficiently. M+ extends MemoryLLM by introducing a co-trained retriever that provides scalable long-term memory; specifically, the retriever extracts and reuses evicted tokens during inference.

Beyond their strategies for updating the memory pool, both MemoryLLM and M+ rely on a series of continual training strategies for knowledge incorporation and forgetting mitigation. The computational overhead of such continual training has limited their publicly released models to relatively small scales (based on Llama-2-7B or Llama-3.1-8B). 
Moreover, their paradigm of concatenating hidden states introduces substantial \textit{positional encoding misalignment}, as the appended states no longer correspond to their original positions.

\section{Supplementary Descriptions for Attention Density}
\subsection{Derivation of Attention Density}
\label{appendix:attention-density-derivation}
This appendix provides the full derivation of the head-averaged
attention matrix $\bar{A}^l$ used in Equation~\ref{eq:density}.

\paragraph{Notation.}
Let $W_{Q,h}^{l} \in \mathbb{R}^{d \times d_k}$ and
$W_{K,h}^{l} \in \mathbb{R}^{d \times d_k}$ be the query and key
projection matrices of layer $l$ for the $h$-th attention head,
where $H$ is the total number of heads and $d_k = d / H$.

\paragraph{Memory as a per-layer prefix.}
The memory is stored as a \emph{per-layer prefix}: at every transformer layer $l$, the stored per-layer hidden states
\begin{equation}
    H_\theta^{l}
    = [\texttt{bos}^{l};\, m_1^{l};\, m_2^{l};\, \dots;\, m_n^{l}]
    \in \mathbb{R}^{(1+N_{n}) \times d}
\end{equation}
are prepended to whatever prompt-side states the layer is currently
processing. The prompt hidden states are therefore produced
\emph{inside} this memory-conditioned forward pass: denoting the prompt-side output of layer $l$ as
$H_p^{l} \in \mathbb{R}^{|p| \times d}$, the full sequence seen by
layer $l$ is
\begin{equation}
    H^{l} = [H_\theta^{l};\, H_p^{l}]
    \in \mathbb{R}^{S \times d},
    \qquad S = 1 + N_n + |p|,
\end{equation}
which is exactly the input used in
Equation~\ref{eq:concat-seq}. Note that $H_p^{l}$ depends on
$H_\theta^{l}$ through the layer's self-attention, so the memory
prefix shapes the prompt representation at every layer.

\paragraph{Per-head attention.}
For each head $h \in \{1, \dots, H\}$, we only need the
attention from the prompt tokens to the full sequence.
We therefore project queries from the prompt hidden states
$H_p^{l}$ and keys from the full sequence $H^{l}$, and apply
scaled dot-product attention under a causal mask
$M_{\mathrm{causal}} \in \mathbb{R}^{|p| \times S}$ that prevents
each prompt position from attending to later prompt positions:
\begin{equation}
\begin{aligned}
    Q_h^l &= H_p^{l}\, W_{Q,h}^l \in \mathbb{R}^{|p| \times d_k}, \\[4pt]
    K_h^l &= H^{l}\, W_{K,h}^l  \in \mathbb{R}^{S \times d_k}, \\[4pt]
    A_h^l &= \mathrm{softmax}\!\left(
        \frac{Q_h^l\, K_h^{l\top}}{\sqrt{d_k}}
        + M_{\mathrm{causal}}
    \right) \in \mathbb{R}^{|p| \times S}.
\end{aligned}
\end{equation}

Equivalently, $A_h^l$ is the last $|p|$ rows of the full
$S \times S$ self-attention matrix, which is all that is needed
for computing $\rho_i^l$.

\paragraph{Head averaging and last-position extraction.}
The per-head attention matrices are averaged across heads,
\begin{equation}
    \bar{A}^l
    = \frac{1}{H} \sum_{h=1}^{H} A_h^l
    \in \mathbb{R}^{|p| \times S},
\end{equation}
and the last-position attention score used in
Equation~\ref{eq:density} is the final row of $\bar{A}^l$
(i.e., the attention emitted by the prompt's last token):
\begin{equation}
\label{eq:last-position-attention}
    \bar{a}_t^l = \bar{A}_{|p|,\,t}^l, \qquad t = 1, \dots, S.
\end{equation}

\subsection{Variants of Attention Computation}
\label{appendix:Variants of Attention Computation}

Equation~\ref{eq:last-position-attention} extracts the attention at the last position. This extends to the mean attention across all prompt tokens:
\begin{equation}
\label{eq:all-position attention}
    \bar{a}_t^{\,l} = \frac{1}{|p|}\sum_{\mathrm{pos}=1}^{|p|} \bar{A}_{\mathrm{pos},\,t}^{\,l}, \quad t = 1,\dots,S.
\end{equation}

\begin{table}[!t]
\centering
\resizebox{0.9\linewidth}{!}{
\begin{tabular}{c|c|ccccc}
\toprule
Layer & $\overline{\text{Rank}}\downarrow$ & Top-$1\uparrow$ & Top-$2\uparrow$ & Top-$3\uparrow$ & Top-$4\uparrow$ & Top-$5\uparrow$ \\
\hline
\rowcolor{blue!8} \multicolumn{7}{c}{\textit{NaturalQA, last-token attention}} \\
\hline
13 & 1.66 & 85.6 & 94.7 & 95.7 & 96.3 & 96.6 \\
9  & 2.14 & 78.3 & 84.9 & 92.2 & 93.2 & 93.8 \\
14 & 3.19 & 75.1 & 83.2 & 85.3 & 86.2 & 86.9 \\
17 & 4.34 & 68.2 & 76.7 & 78.1 & 79.2 & 79.9 \\
11 & 5.01 & 60.7 & 72.3 & 73.9 & 74.7 & 75.5 \\
\hline
\rowcolor{blue!8} \multicolumn{7}{c}{\textit{NaturalQA, all-token attention}} \\
\hline
13 & 2.82 & 58.7 & 86.1 & 88.5 & 89.4 & 90.3 \\
9  & 3.68 & 38.0 & 72.1 & 81.0 & 83.5 & 84.9 \\
11 & 4.38 & 37.9 & 74.8 & 77.8 & 79.7 & 80.8 \\
8  & 4.71 & 40.5 & 69.3 & 74.6 & 77.2 & 78.5 \\
14 & 4.86 & 53.0 & 75.2 & 77.3 & 78.4 & 79.1 \\
\hline
\rowcolor{blue!8} \multicolumn{7}{c}{\textit{SQuAD, last-token attention}} \\
\hline
13 & 2.61 & 74.8 & 82.1 & 85.9 & 85.9 & 89.2 \\
14 & 3.60 & 69.3 & 76.6 & 80.1 & 82.5 & 83.8 \\
17 & 3.74 & 75.1 & 81.0 & 82.5 & 83.6 & 84.2 \\
30 & 4.81 & 53.9 & 66.7 & 71.5 & 74.0 & 76.1 \\
20 & 5.21 & 69.1 & 73.0 & 74.6 & 75.6 & 76.3 \\
\hline
\rowcolor{blue!8} \multicolumn{7}{c}{\textit{SQuAD, all-token attention}} \\
\hline
13 & 1.90 & 80.2 & 86.1 & 91.8 & 93.1 & 94.0 \\
14 & 2.80 & 75.4 & 82.7 & 86.7 & 88.0 & 89.0 \\
9  & 3.22 & 43.9 & 62.8 & 79.2 & 85.7 & 87.7 \\
8  & 3.66 & 45.1 & 68.1 & 74.4 & 82.5 & 84.5 \\
10 & 4.18 & 44.6 & 67.0 & 76.0 & 80.6 & 82.3 \\
\bottomrule
\end{tabular}}
\caption{Layers with the highest tracing capability based on 32-layer Llama-3.1-8B-Instruct, ranked by average attention density rank, with Top-$K$ accuracies (\%) across all positions. Results are shown for two datasets (NaturalQA, SQuAD) under two attention computation strategies (last-token and all-token).}
\label{tab:retrieval_layers_combined_Llama-3.1-8B-Instruct}
\end{table}

\begin{table}[!t]
\centering
\resizebox{0.9\linewidth}{!}{
\begin{tabular}{c|c|ccccc}
\toprule
Layer & $\overline{\text{Rank}}\downarrow$ & Top-$1\uparrow$ & Top-$2\uparrow$ & Top-$3\uparrow$ & Top-$4\uparrow$ & Top-$5\uparrow$ \\
\hline
\rowcolor{blue!8} \multicolumn{7}{c}{\textit{NaturalQA, last-token attention}} \\
\hline
27 & 3.38 & 56.9 & 75.5 & 79.9 & 83.0 & 84.7 \\
26 & 4.81 & 61.2 & 68.0 & 71.2 & 73.2 & 75.1 \\
2 & 5.06 & 18.6 & 49.6 & 51.3 & 61.8 & 63.2 \\
3 & 5.12 & 8.2 & 27.9 & 58.5 & 67.8 & 74.2 \\
13 & 6.44 & 37.2 & 57.2 & 63.1 & 67.7 & 69.2 \\
\hline
\rowcolor{blue!8} \multicolumn{7}{c}{\textit{NaturalQA, all-token attention}} \\
\hline
14 & 2.02 & 86.9 & 90.3 & 92.1 & 92.9 & 93.7 \\
19 & 2.97 & 75.0 & 84.7 & 86.7 & 87.7 & 88.4 \\
16 & 3.00 & 77.4 & 85.4 & 86.8 & 87.9 & 88.6 \\
27 & 3.04 & 60.4 & 76.4 & 83.2 & 85.7 & 87.2 \\
22 & 4.27 & 73.3 & 77.1 & 79.0 & 80.0 & 80.7 \\
\hline
\rowcolor{blue!8} \multicolumn{7}{c}{\textit{SQuAD, last-token attention}} \\
\hline
27 & 2.60 & 60.6 & 73.8 & 80.4 & 85.5 & 88.7 \\
26 & 3.61 & 61.7 & 70.0 & 74.5 & 79.0 & 81.7 \\
25 & 5.29 & 55.7 & 63.5 & 68.7 & 71.7 & 74.1 \\
2 & 5.31 & 12.6 & 27.7 & 30.8 & 53.5 & 55.3 \\
3 & 6.20 & 5.3 & 10.7 & 38.9 & 51.5 & 60.6 \\
\hline
\rowcolor{blue!8} \multicolumn{7}{c}{\textit{SQuAD, all-token attention}} \\
\hline
14 & 1.45 & 92.5 & 95.2 & 96.2 & 97.0 & 97.4 \\
19 & 1.64 & 85.4 & 93.7 & 95.3 & 96.3 & 96.8 \\
27 & 1.86 & 69.7 & 85.9 & 90.6 & 93.3 & 95.0 \\
22 & 1.89 & 87.0 & 91.2 & 92.8 & 93.9 & 94.5 \\
16 & 1.97 & 85.0 & 90.9 & 93.1 & 94.2 & 94.5 \\
\bottomrule
\end{tabular}}
\caption{Layers with the highest tracing capability based on 28-layer Qwen2.5-7B-Instruct, ranked by average attention density rank, with Top-$K$ accuracies (\%) across all positions. Results are shown for two datasets (NaturalQA, SQuAD) under two attention computation strategies (last-token and all-token).}
\label{tab:retrieval_layers_combined_Qwen2.5-7B-Instruct}
\end{table}

\begin{table}[!t]
\centering
\resizebox{0.9\linewidth}{!}{
\begin{tabular}{c|c|ccccc}
\toprule
Layer & $\overline{\text{Rank}}\downarrow$ & Top-$1\uparrow$ & Top-$2\uparrow$ & Top-$3\uparrow$ & Top-$4\uparrow$ & Top-$5\uparrow$ \\
\hline
\rowcolor{blue!8} \multicolumn{7}{c}{\textit{NaturalQA, last-token attention}} \\
\hline
16 & 1.80 & 80.0 & 88.3 & 91.3 & 93.6 & 94.9 \\
15 & 2.00 & 82.4 & 89.2 & 91.6 & 93.0 & 93.8 \\
18 & 2.28 & 76.1 & 87.0 & 89.1 & 90.3 & 91.1 \\
19 & 2.53 & 72.3 & 84.3 & 87.1 & 89.3 & 90.4 \\
14 & 3.51 & 65.1 & 72.8 & 79.6 & 81.4 & 83.0 \\
\hline
\rowcolor{blue!8} \multicolumn{7}{c}{\textit{NaturalQA, all-token attention}} \\
\hline
15 & 1.70 & 86.2 & 93.4 & 94.4 & 95.0 & 95.4 \\
16 & 2.32 & 75.2 & 88.4 & 89.7 & 90.6 & 91.4 \\
11 & 2.42 & 54.4 & 88.3 & 90.3 & 92.0 & 92.9 \\
18 & 2.76 & 73.4 & 85.8 & 87.3 & 88.5 & 89.5 \\
12 & 2.79 & 55.7 & 88.9 & 90.2 & 90.7 & 91.2 \\
\hline
\rowcolor{blue!8} \multicolumn{7}{c}{\textit{SQuAD, last-token attention}} \\
\hline
16 & 1.40 & 87.9 & 92.8 & 95.3 & 96.8 & 97.5 \\
18 & 1.41 & 89.9 & 93.8 & 95.7 & 96.8 & 97.6 \\
15 & 1.45 & 88.8 & 92.9 & 95.2 & 96.5 & 97.2 \\
19 & 1.74 & 84.5 & 90.5 & 93.0 & 94.4 & 95.3 \\
17 & 3.27 & 64.4 & 76.9 & 81.2 & 83.8 & 85.4 \\
\hline
\rowcolor{blue!8} \multicolumn{7}{c}{\textit{SQuAD, all-token attention}} \\
\hline
15 & 1.22 & 94.0 & 96.6 & 97.4 & 97.9 & 98.3 \\
12 & 1.55 & 81.8 & 92.4 & 95.0 & 96.3 & 97.0 \\
16 & 1.60 & 87.3 & 92.4 & 94.4 & 95.5 & 96.4 \\
18 & 1.63 & 87.3 & 93.0 & 94.8 & 95.6 & 96.1 \\
9 & 1.80 & 75.6 & 88.8 & 92.8 & 94.6 & 95.6 \\
\bottomrule
\end{tabular}}
\caption{Layers with the highest tracing capability based on 32-layer Mistral-7B-Instruct-v0.3, ranked by average attention density rank, with Top-$K$ accuracies (\%) across all positions. Results are shown for two datasets (NaturalQA, SQuAD) under two attention computation strategies (last-token and all-token).}
\label{tab:retrieval_layers_combined_Mistral-7B-Instruct-v0.3}
\end{table}

\begin{table}[!t]
\centering
\resizebox{0.9\linewidth}{!}{
\begin{tabular}{c|c|ccccc}
\toprule
Layer & $\overline{\text{Rank}}\downarrow$ & Top-$1\uparrow$ & Top-$2\uparrow$ & Top-$3\uparrow$ & Top-$4\uparrow$ & Top-$5\uparrow$ \\
\hline
\rowcolor{blue!8} \multicolumn{7}{c}{\textit{NaturalQA, last-token attention}} \\
\hline
15 & 1.24 & 78.6 & 97.8 & 99.5 & 99.9 & 99.9 \\
17 & 1.43 & 61.9 & 95.6 & 99.7 & 100.0 & 100.0 \\
9 & 1.73 & 32.4 & 94.6 & 99.9 & 100.0 & 100.0 \\
19 & 2.08 & 31.2 & 75.4 & 94.4 & 97.2 & 98.3 \\
13 & 2.68 & 20.3 & 44.5 & 79.7 & 90.8 & 97.4 \\
\hline
\rowcolor{blue!8} \multicolumn{7}{c}{\textit{NaturalQA, all-token attention}} \\
\hline
15 & 1.25 & 79.1 & 96.9 & 99.2 & 100.0 & 100.0 \\
17 & 1.46 & 61.4 & 93.7 & 99.2 & 99.6 & 100.0 \\
9 & 1.66 & 37.0 & 97.5 & 99.6 & 100.0 & 100.0 \\
19 & 2.21 & 26.5 & 68.7 & 95.1 & 96.7 & 98.1 \\
7 & 2.24 & 11.2 & 73.6 & 93.2 & 98.8 & 99.7 \\
\hline
\rowcolor{blue!8} \multicolumn{7}{c}{\textit{SQuAD, last-token attention}} \\
\hline
15 & 1.44 & 61.7 & 95.7 & 99.4 & 100.0 & 100.0 \\
17 & 1.46 & 54.5 & 99.0 & 99.9 & 99.9 & 99.9 \\
9 & 1.97 & 9.4 & 93.9 & 99.0 & 99.9 & 99.9 \\
19 & 2.39 & 18.8 & 65.9 & 88.9 & 93.3 & 96.5 \\
23 & 2.52 & 60.4 & 75.4 & 81.3 & 84.8 & 87.7 \\
\hline
\rowcolor{blue!8} \multicolumn{7}{c}{\textit{SQuAD, all-token attention}} \\
\hline
15 & 1.27 & 73.9 & 99.3 & 100.0 & 100.0 & 100.0 \\
17 & 1.41 & 59.3 & 99.9 & 100.0 & 100.0 & 100.0 \\
9 & 1.83 & 17.2 & 99.4 & 100.0 & 100.0 & 100.0 \\
23 & 1.98 & 65.4 & 82.2 & 86.9 & 91.0 & 93.1 \\
13 & 2.13 & 13.3 & 75.6 & 98.7 & 99.8 & 99.9 \\
\bottomrule
\end{tabular}}
\caption{Layers with the highest tracing capability based on 42-layer Gemma-2-9b-it, ranked by average attention density rank, with Top-$K$ accuracies (\%) across all positions. Results are shown for two datasets (NaturalQA, SQuAD) under two attention computation strategies (last-token and all-token).}
\label{tab:retrieval_layers_combined_Gemma-2-9b-it}
\end{table}

Correspondingly, we extend the investigation of attention density as a tracing signal (Section~\ref{sec: Attention Density as Retrieval Signal} and Table~\ref{tab:retrieval_layers}) (originally based on NaturalQA with last-token attention) to additional configurations: NaturalQA with all-token attention, and SQuAD with both last-token and all-token attention. The results are reported in Table~\ref{tab:retrieval_layers_combined_Llama-3.1-8B-Instruct}.
Layer~13 consistently exhibits the highest tracing capacity across all configurations. 
In this experiment set, last-token attention is better suited to NaturalQA, while all-token attention is better suited to SQuAD.

\section{Results and Analyses in Settings without Positional Distortion}
\subsection{Analyses of Attention Density without Positional Distortion}
\label{app:Analyses of Attention Density without Positional Distortion}

\begin{figure}[!t]
    \centering
    \begin{subfigure}[b]{0.8\linewidth}
        \centering
        \includegraphics[width=\textwidth]{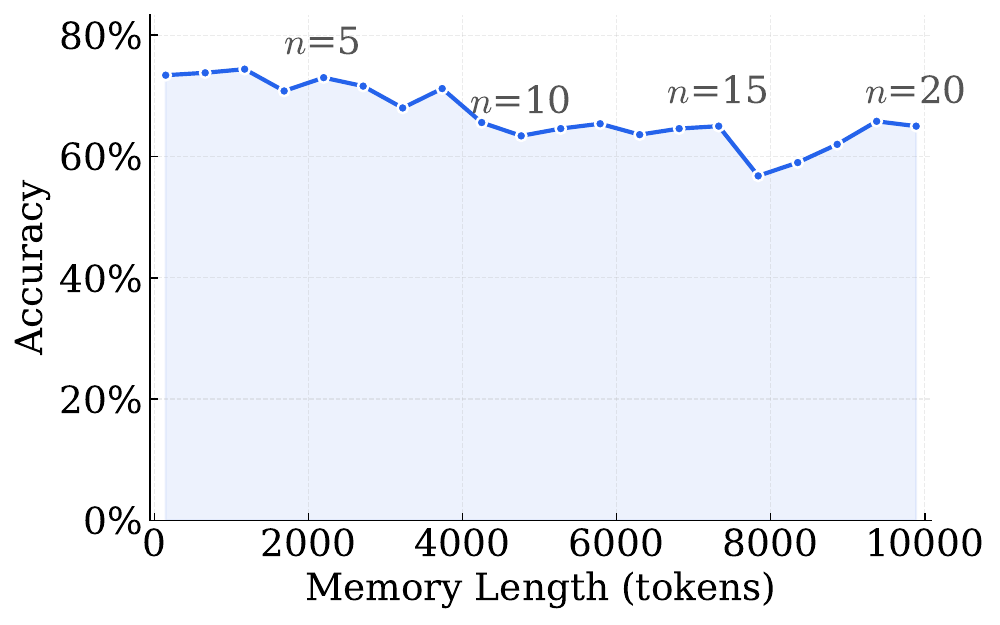}
        \caption{QA accuracy decreasing as the memory length (in tokens) increases.}
        \label{fig:investigation_sub1_no_distortion}
    \end{subfigure}
    \hfill
    \begin{subfigure}[b]{0.8\linewidth}
        \centering
        \includegraphics[width=\textwidth]{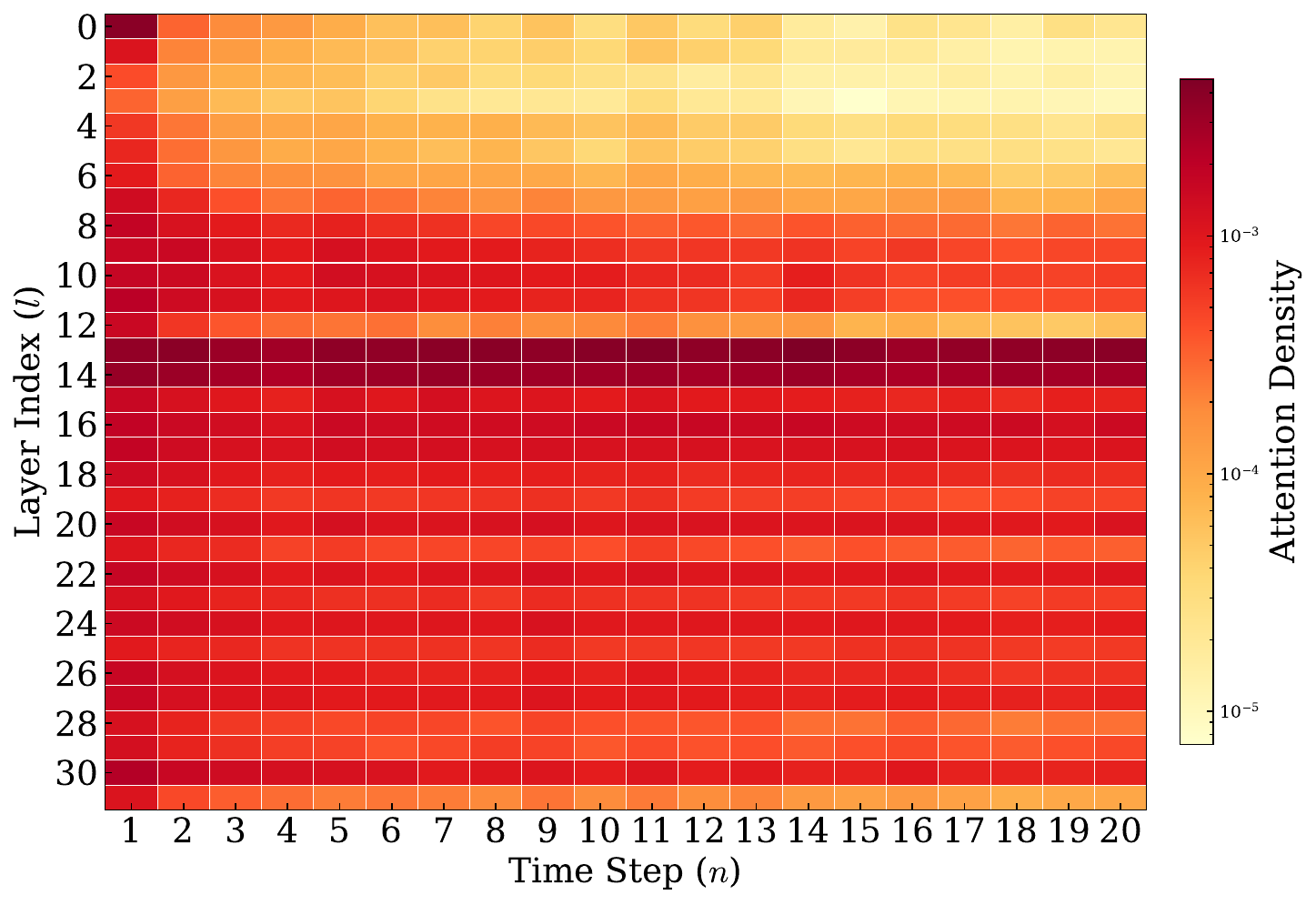}
        \caption{Attention density $\rho_1^l$ of the target knowledge fragment $x_1$ across layers $l$ and time steps $n$.}
        \label{fig:investigation_sub2_no_distortion}
    \end{subfigure}
    \hfill
        \begin{subfigure}[b]{0.8\linewidth}
        \centering
        \includegraphics[width=\textwidth]{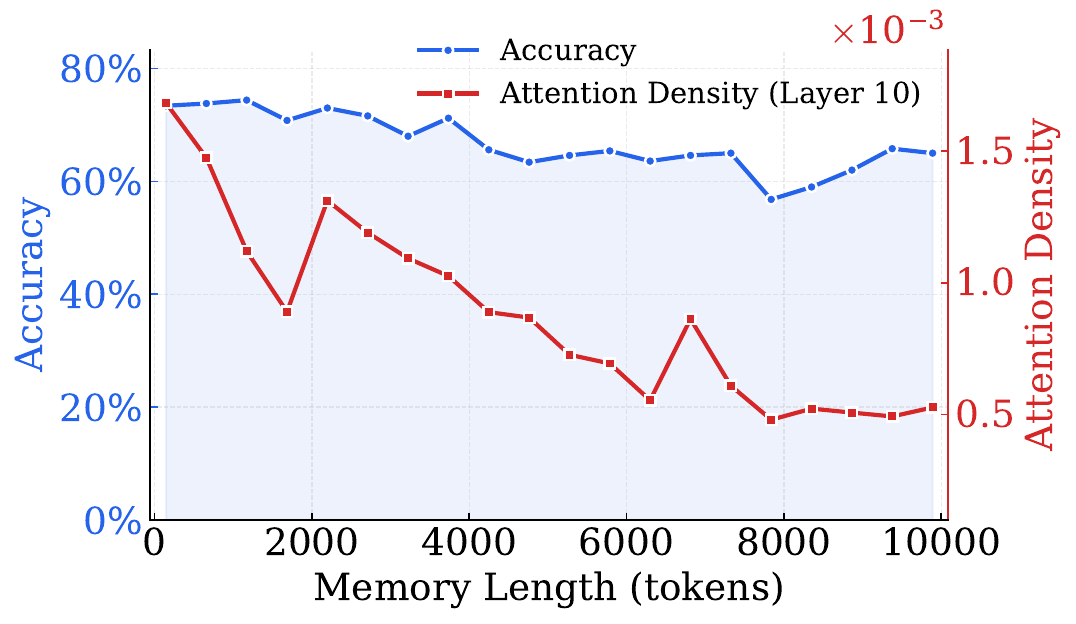}
        \caption{QA accuracy and attention density $\rho_1^{10}$ of the target knowledge fragment at layer 10.}
        \label{fig:investigation_sub4_no_distortion}
    \end{subfigure}
    \hfill
    \begin{subfigure}[b]{\linewidth}
        \centering
        \includegraphics[width=\textwidth]{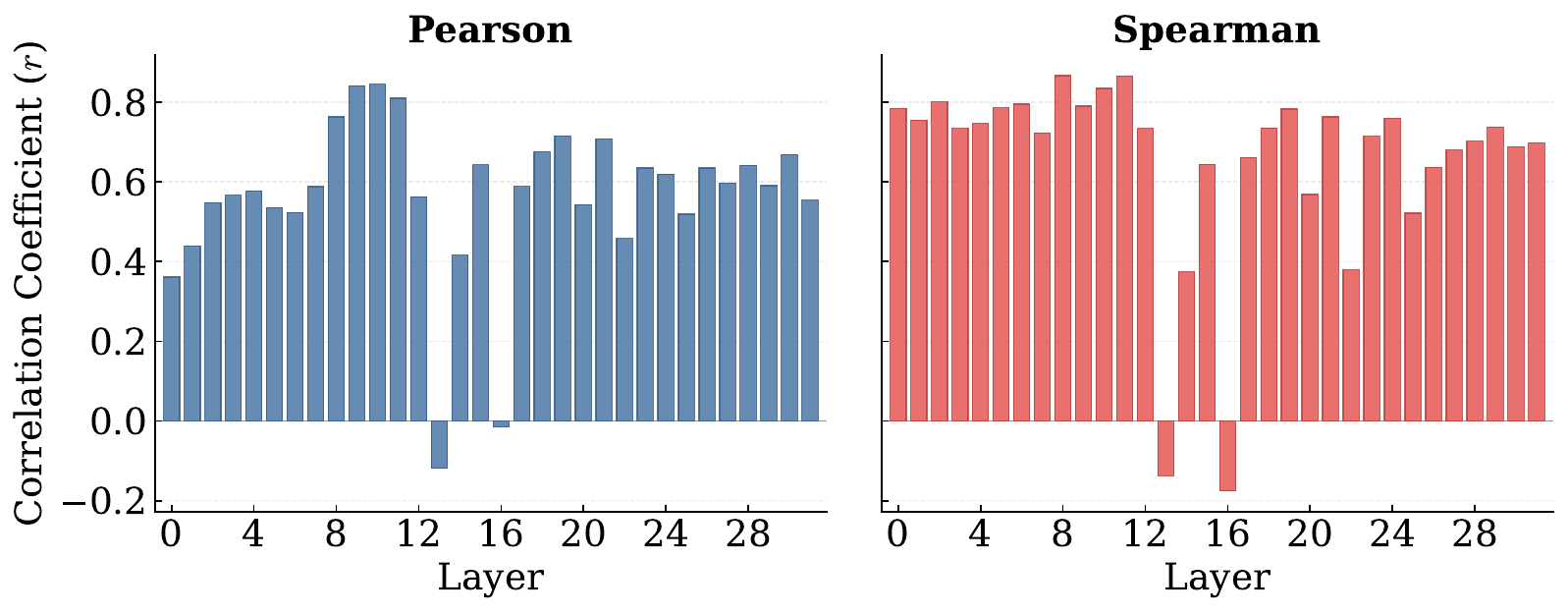}
        \caption{Per-layer Pearson and Spearman correlation coefficients between attention density $\rho_i^l$ and QA accuracy across 20 steps.}
        \label{fig:investigation_sub3_no_distortion}
    \end{subfigure}
    
    \caption{Effects and analyses attention density without positional distortion.}
    \label{fig:three_subfigures_no_distortion}
\end{figure}

This section presents a mirrored version of the experiments in Section~\ref{sec: Effects of Positional Distortion} and Section~\ref{sec: investigation_attention}, where the latent-memory paradigm is removed and all updates are performed over text tokens. As a result, positional encodings are correctly assigned throughout the input sequence.
Figure~\ref{fig:three_subfigures_no_distortion} reports the corresponding results on Llama-3.1-8B-Instruct and NaturalQA. Comparing Figure~\ref{fig:three_subfigures_no_distortion} with the positional-distortion results in Figure~\ref{fig:three_subfigures}, we make two observations.

(1) In settings without positional distortion, accuracy remains more stable as the number of update steps increases, even though layer 10, which exhibits the highest correlation with accuracy, shows a substantial drop.

(2) As shown in Figure~\ref{fig:investigation_sub3_no_distortion}, the layer-wise Pearson and Spearman correlation coefficients between attention density and QA accuracy are generally lower in settings without positional distortion.

These observations suggest that, when positional distortion is absent, attention density is no longer as significant an indicator of performance as it is when positional distortion is present.

\subsection{Analyses of Tracing Capacities without Positional Distortion}
\label{app:Analyses of Tracing Capacities without Positional Distortion}

This section presents a mirrored version of the experiments in Section~\ref{sec: Attention Density as Retrieval Signal}, where the latent-memory paradigm is removed and the concatenated input is entirely text tokens. Thus, positional encodings are correctly assigned throughout the input sequence.
Table~\ref{tab:retrieval_layers_no_distortion} reports the average attention density rank with Top-$K$ accuracy of the target fragment across all positions for all layers. We make the following observations:

(1) Layer 13 exhibits the highest tracing capability, which is consistent with the results observed under positional distortion in Table~\ref{tab:retrieval_layers}. This result is also implied by the high attention density of layer 13 of the target fragment shown in Figure~\ref{fig:investigation_sub2_no_distortion}.

(2) Overall, layer-wise tracing capabilities are substantially higher in settings without positional distortion than in positional-distortion settings. This suggests that positional distortion can compromise the ability of attention density to trace relevant information across layers, even though this tracing capability remains effective under the latent-memory paradigm, as discussed in Section~\ref{sec: Attention Density as Retrieval Signal}.

(3) As shown by the layer distribution in Table~\ref{tab:retrieval_layers_no_distortion}, intermediate layers exhibit much higher tracing capabilities than earlier or final layers. This observation is consistent with the explanation proposed in Section~\ref{sec: Attention Density as Retrieval Signal}: Intermediate layers can encode richer representations, which benefits feature relevance and extraction~\cite{pmlr-v267-skean25a}.

(4) An interesting finding stands out: Tracing capabilities can vary substantially between neighboring layers (especially the gap observed between layer 12 and layer 13). This sharp transition and the general varying tendency suggest that tracing capabilities do not evolve smoothly across depth.


\begin{table}[!t]
\centering
\resizebox{\linewidth}{!}{
\begin{tabular}{c|c|ccccc}
\toprule
Layer & $\overline{\text{Rank}}\downarrow$ & Top-$1\uparrow$ & Top-$2\uparrow$ & Top-$3\uparrow$ & Top-$4\uparrow$ & Top-$5\uparrow$ \\
\midrule
13 & 1.00 & 100.0\% & 100.0\% & 100.0\% & 100.0\% & 100.0\% \\
14 & 1.00 & 99.9\% & 100.0\% & 100.0\% & 100.0\% & 100.0\% \\
17 & 1.00 & 99.9\% & 100.0\% & 100.0\% & 100.0\% & 100.0\% \\
15 & 1.00 & 99.5\% & 99.9\% & 99.9\% & 99.9\% & 99.9\% \\
16 & 1.00 & 99.7\% & 100.0\% & 100.0\% & 100.0\% & 100.0\% \\
18 & 1.01 & 99.2\% & 99.7\% & 99.8\% & 99.9\% & 99.9\% \\
19 & 1.01 & 98.8\% & 99.7\% & 99.8\% & 99.9\% & 99.9\% \\
21 & 1.01 & 98.4\% & 99.7\% & 99.8\% & 99.8\% & 99.8\% \\
26 & 1.01 & 98.9\% & 99.7\% & 99.8\% & 99.8\% & 99.8\% \\
20 & 1.02 & 98.7\% & 99.9\% & 99.9\% & 99.9\% & 100.0\% \\
27 & 1.02 & 99.0\% & 99.7\% & 99.9\% & 100.0\% & 100.0\% \\
25 & 1.02 & 98.4\% & 99.5\% & 99.7\% & 99.8\% & 99.9\% \\
29 & 1.02 & 98.9\% & 99.6\% & 99.8\% & 99.9\% & 100.0\% \\
24 & 1.03 & 98.2\% & 99.4\% & 99.7\% & 99.7\% & 99.8\% \\
30 & 1.03 & 98.8\% & 99.6\% & 99.7\% & 99.8\% & 99.9\% \\
9 & 1.03 & 98.2\% & 99.3\% & 99.6\% & 99.7\% & 99.8\% \\
22 & 1.03 & 98.0\% & 99.2\% & 99.7\% & 99.8\% & 99.9\% \\
23 & 1.04 & 97.4\% & 99.3\% & 99.6\% & 99.7\% & 99.8\% \\
10 & 1.04 & 98.3\% & 99.1\% & 99.5\% & 99.6\% & 99.7\% \\
28 & 1.09 & 95.6\% & 98.0\% & 98.8\% & 99.1\% & 99.3\% \\
11 & 1.13 & 92.0\% & 97.8\% & 98.8\% & 99.3\% & 99.6\% \\
8 & 1.29 & 87.6\% & 94.0\% & 97.1\% & 98.3\% & 98.8\% \\
31 & 1.88 & 54.4\% & 89.9\% & 93.0\% & 94.7\% & 95.9\% \\
7 & 1.94 & 62.3\% & 84.8\% & 87.3\% & 89.9\% & 96.6\% \\
6 & 2.22 & 34.6\% & 80.1\% & 81.3\% & 96.5\% & 97.1\% \\
3 & 2.83 & 32.1\% & 66.1\% & 71.3\% & 84.7\% & 90.4\% \\
4 & 2.96 & 19.6\% & 57.8\% & 72.0\% & 90.8\% & 91.9\% \\
5 & 3.52 & 12.1\% & 51.8\% & 55.1\% & 77.6\% & 89.1\% \\
12 & 3.83 & 39.7\% & 64.2\% & 71.8\% & 75.8\% & 79.1\% \\
2 & 4.89 & 9.1\% & 41.4\% & 45.4\% & 55.4\% & 68.5\% \\
1 & 7.11 & 5.2\% & 24.4\% & 30.9\% & 35.6\% & 53.2\% \\
0 & 13.34 & 5.0\% & 8.5\% & 11.0\% & 17.1\% & 19.8\% \\
\bottomrule
\end{tabular}}
\caption{Layer-wise tracing capability (\textit{free from positional distortion}) based on 32-layer Llama-3.1-8B-Instruct, NaturalQA and last-token attention, ranked by average attention density rank, with Top-$K$ accuracies (\%) across all positions.}
\label{tab:retrieval_layers_no_distortion}
\end{table}

\begin{algorithm}[t]
\small
\caption{Memory Evolution with Forgetting}
\label{alg:memory_evolution}
\begin{algorithmic}[1]
\Input Current memory $\theta_{n-1} = \{m_1,\dots,m_{n-1}\}$ with retained lengths $\{L_1,\dots,L_{n-1}\}$, new knowledge $x_{n}$, maximum capacity $N_{\max}$
\Output Updated memory $\theta_{n}$

\State $m_{n} \gets f_{\mathrm{form}}(x_{n})$ \Comment{Memory formation}
\State $L_{n} \gets |x_{n}|$
\State $N_{n-1} \gets \sum_{i=1}^{n-1} L_i$ \Comment{Current total memory length}

\If{$N_{n-1} + L_{n} \leq N_{\max}$}
    \State $\theta_{n} \gets \text{concat}(\theta_{n-1}, m_{n})$
\Else
    \State $N_{\text{forget}} \gets N_{n-1} + L_{n} - N_{\max}$
    \State $\theta_{n-1} \gets f_{\mathrm{forget}}(\theta_{n-1},\; N_{\text{forget}})$ \Comment{Alg.~\ref{alg:proportional_forgetting}}
    \State $\theta_{n} \gets \text{concat}(\theta_{n-1}, m_{n})$
\EndIf

\State \Return $\theta_{n}$
\end{algorithmic}
\end{algorithm}

\begin{algorithm}[t]
\small
\caption{Proportional Forgetting Based on Self-Information}
\label{alg:proportional_forgetting}
\begin{algorithmic}[1]
\Input Memory $\theta_{n-1} = \{m_1,\dots,m_{n-1}\}$, retained lengths $\{L_1,\dots,L_{n-1}\}$, precomputed self-information scores $\{I(x_i^{(t)})\}$ for each knowledge $i$ and position $t$, forgetting budget $N_{\mathrm{forget}}$
\Output Pruned memory $\theta_{n-1}$ with updated lengths

\State $N_{n-1} \gets \sum_{i=1}^{n-1} L_i$

\For{$i = 1$ \textbf{to} $n-1$} \Comment{Proportional quota allocation}
    \State $N_{\mathrm{forget},i} \gets \left\lfloor N_{\mathrm{forget}} \cdot \dfrac{L_i}{N_{n-1}} \right\rfloor$
\EndFor
\State Distribute remaining $N_{\mathrm{forget}} - \sum_i N_{\mathrm{forget},i}$ to entries with the largest fractional remainders

\For{$i = 1$ \textbf{to} $n-1$} \Comment{Pruning}
    \If{$N_{\mathrm{forget},i} > 0$}
        \State $D_i \gets \operatorname{argtopk}_{t \in \{1,\dots,L_i\}}\big(-I(x_i^{(t)}),\; N_{\mathrm{forget},i}\big)$
        \For{each layer $l = 1, \dots, L$}
            \State $m_i^l \gets m_i^l\big[\{1,\dots,L_i\} \setminus D_i,\; :\big]$
        \EndFor
        \State $L_i \gets L_i - N_{\mathrm{forget},i}$
    \EndIf
\EndFor

\State \Return $\theta_{n-1}$
\end{algorithmic}
\end{algorithm}

\section{Descriptions of Informativeness-Based Proportional Forgetting}
\label{app:Descriptions of Informativeness-Based Proportional Forgetting}

Let $L_i \ (i = 1,\dots,n-1)$  denote the \textbf{current} number of hidden states retained for knowledge $x_i$ (initially $L_i = |x_i|$; this quantity decreases monotonically as forgetting is applied). For each retained position 
$t$ in the current memory, we measure the self-information of the corresponding token as:
\begin{equation}
I(x_i^{(t)}) = -\log p_\phi(x_i^{(t)} \mid x_i^{(<t)}), 
\end{equation} 
where $p_\phi(\cdot)$ denotes the token probability estimated by the language model. Tokens with lower self-information are deemed more redundant. Given a forgetting quota of $N_{\text{forget},i}$ for knowledge $x_i$, we select the $N_{\text{forget},i}$ positions with the lowest self-information for removal:
\begin{equation}
D_i = \operatorname{argtopk}_{t \in {1,\dots,L_i}} \big(-I(x_i^{(t)}), N_{\text{forget},i}\big), 
\end{equation}
where $D_i$  is the index set of positions to be dropped, and $\operatorname{argtopk}(\cdot,k)$ returns the indices of the $k$ largest values. The pruned memory at layer $l$ is then obtained by removing these positions:
\begin{equation}
m_i^{l} \leftarrow m_i^l\big[{1, \dots, L_i} \setminus {D}_i; :\big] \in \mathbb{R}^{(L_i - N_{\text{forget},i}) \times d}, \end{equation} 
and the retained length is updated accordingly: 
\begin{equation} 
L_i \leftarrow L_i - N_{\text{forget},i}. 
\end{equation} 
This ensures that the most informative hidden states are preferentially retained under a fixed capacity budget, while the retained length of each knowledge fragment shrinks over successive evolution steps.
Algorithm~\ref{alg:proportional_forgetting} details the process of proportional forgetting.


\begin{figure}[!t]
    \centering
    \includegraphics[width=\linewidth]{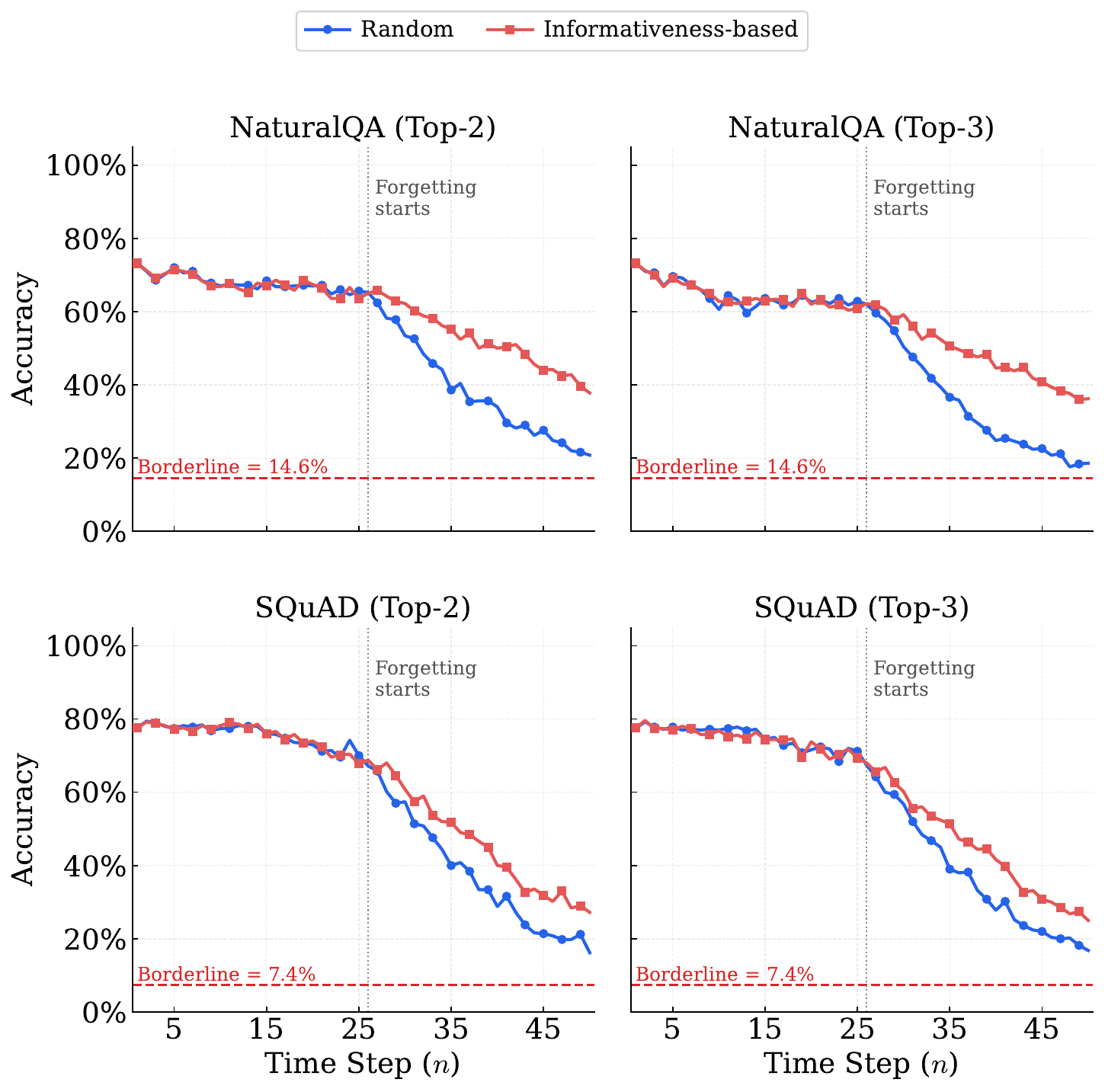}
    \caption{50-step knowledge retention comparing informativeness-based vs. random forgetting on NaturalQA and SQuAD, of which Top-2 or Top-3 highest-ranked memory fragments are filtered of defragmentation.}
    \label{fig:retention_different_forgetting}
\end{figure}

\section{Supplementary Analyses and Descriptions for Experiments}
\subsection{Supplementary Results of Ablation Study of Forgetting Strategies}
\label{app:Supplementary Results of Ablation Study of Forgetting Strategies}

The maximum time step is 50, which allows the memory length $N_i$ (when $i \geq 26$) to reach its maximum length $N_{\text{max}}$ and triggers the forgetting strategy after time step 26.
Both Top-2 and Top-3 defragmentation settings are evaluated on NaturalQA and SQuAD. Figure~\ref{fig:retention_different_forgetting} presents the results, from which we find that: The advantage of informativeness-based forgetting is validated and their gap increases over time. Specifically, on NaturalQA with Top-2, informativeness-based forgetting sustains an accuracy of 37.8\% at step 50, compared to 20.8\% for random forgetting, indicating an improvement of 81.73\%.

\subsection{Setups for Long Context Experiments}
\label{app:Setups for Long Context Experiments}
As MemoryLLM and M+ adopt a fixed (512 to 256) compression strategy, for each long context, it is split into 512-token chunks for memory injection.
For MemDefrag which can take advantage of flexible chunking size, for the six datasets, there are two chunking strategies:

(1) Passage-based chunking: For multi-doc QA datasets where the context is explicitly structured into labeled passages (\textit{2WikiMultihopQA}, \textit{HotpotQA}, and \textit{MuSiQue}), the context is split at each delimiter of \textit{``Passage N''}, treating each passage as a natural chunk. This preserves the inherent document boundaries provided by the dataset.

(2) Token-based chunking: For single-doc QA datasets where the context is a continuous document without explicit passage markers (\textit{MultiFieldQA-en}, \textit{NarrativeQA}, and \textit{Qasper}), the context is split into chunks of at least 256 tokens each, breaking at sentence boundaries (periods or newlines). If the last chunk is shorter than 256 tokens, it is merged into the preceding chunk to avoid overly short trailing segments.

\begin{table*}[t] 
\centering
\resizebox{\textwidth}{!}{%
\begin{tabular}{l ccc ccccccc}
\toprule
Method   & Vanilla & MemoryLLM & M+ & Top-1 & Top-2 & Top-3 & Top-4 & Top-5 & Top-6 & Top-all \\
\cmidrule(lr){2-4} \cmidrule(lr){5-11}
(NaturalQA) Time (s) & 59.30      & 64.19        & 87.24      & 75.07    & 76.49    & 78.96    & 82.15    & 86.81    & 90.53    & 103.67      \\
(SQuAD) Time (s) & 60.11      & 65.71        & 88.98      & 79.23    &   81.01  & 82.38    & 84.44    & 87.62    & 92.20    & 107.95      \\
\bottomrule
\end{tabular}}
\caption{Average time per experimental group. Top-$K$ refers to Top-$K$ filtering variant of our MemDefrag.} 
\label{tab:runtime}
\end{table*}

\subsection{Efficiency Analyses}
\label{app:Efficiency Analyses}

For the two datasets, NaturalQA and SQuAD, respectively, we list the efficiency of various methods in Table~\ref{tab:runtime}, including vanilla latent memory, MemoryLLM, M+, and several Top-$K$ variants of our MemDefrag. From this table, we observe that:

(1) The running time of MemDefrag generally increases as $K$ becomes larger. This is expected, since a larger $K$ requires the model to keep more candidate memory fragments at each step, thereby introducing additional computational overhead for inference.

(2) When $K \in \{1,2,3,4\}$, MemDefrag not only outperforms MemoryLLM and M+ in terms of accuracy, as shown in Figure~\ref{fig:retention_comparison}, but also achieves higher efficiency than M+. Thus, this range of $K$ provides a favorable trade-off between effectiveness and efficiency.

\paragraph{GPU memory consumption.}
We further analyze the peak GPU memory consumption of MemDefrag. The dominant resident cost comes from the model weights and the stored latent memory, whose size is proportional to $N_{\max} \times d \times L$. This memory capacity is identical by construction to those of MemoryLLM and M+, with $N_{\max}=12{,}800$ as specified in Section~\ref{sec: Experiments}.

In addition to this resident cost, memory defragmentation introduces
only transient buffers, primarily including (1) the head-averaged
attention scores computed at the tracer layer and discarded immediately
after ranking, and (2) $O(n)$ integer indices for the permutation
$\pi$. In particular, MemDefrag does not materialize a second
full-size copy of the stored memory; only the selected Top-$K$ states
are gathered at each layer to construct $\theta_n^p$.

We measure peak GPU memory under the same knowledge-retention setting:
Llama-3.1-8B-Instruct on NaturalQA at step $n=50$, where
$N_n=N_{\max}=12{,}800$. The observed variance is zero because the tensor shapes are independent of the evaluation examples.

\begin{table*}[t]
    \centering
    \small
    \setlength{\tabcolsep}{6pt}
    \renewcommand{\arraystretch}{1.15}
    \begin{tabular}{@{}l p{0.55\textwidth} r@{}}
        \toprule
        Path
        & Computation included
        & Peak GPU memory (GB) \\
        \midrule
        Resident baseline
        & Model weights and stored latent memory
        & 18.11 \\
        
        Vanilla
        & Generation with the full 12,800-state latent memory
        & 23.03 \\
        
        \textbf{MemDefrag (Top-2)}
        & Tracing, memory defragmentation, and generation with
          the selected Top-2 memory fragments
        & \textbf{19.73} \\
        \bottomrule
    \end{tabular}
    \caption{Peak GPU memory consumption under the 50-step
    knowledge-retention setting.}
    \label{tab:gpu-memory}
\end{table*}

Table~\ref{tab:gpu-memory} supports two observations.
(1) \textbf{MemDefrag introduces little transient memory overhead
and reaches a lower peak than vanilla latent memory.}
Relative to the 18.11-GB resident baseline, MemDefrag requires only
1.62 GB of additional working memory, compared with 4.92 GB for
vanilla inference. Consequently, MemDefrag reduces peak GPU memory
by 3.30 GB, or approximately 14\%, relative to vanilla inference.
The tracing pass terminates at the tracer layer $l^*$, while the
subsequent generation pass operates on the truncated Top-$K$ memory,
for which $N_K \ll N_n$. Its peak is therefore lower than that of a
generation pass attending to all 12,800 stored states at every layer.

(2) \textbf{MemDefrag also exhibits a competitive memory footprint
relative to existing latent-memory methods.}
For reference, the M+ paper reports GPU memory consumption of
21,176.24 MB for MemoryLLM and 21,177.76 MB for M+
($\approx$ 20.7 GB) under its measurement setting. MemDefrag reaches a lower peak of 19.73 GB while requiring neither a co-trained retriever nor any auxiliary module. 

\begin{figure}[!t]
    \centering
    \begin{subfigure}[b]{\linewidth}
        \centering
        \includegraphics[width=\textwidth]{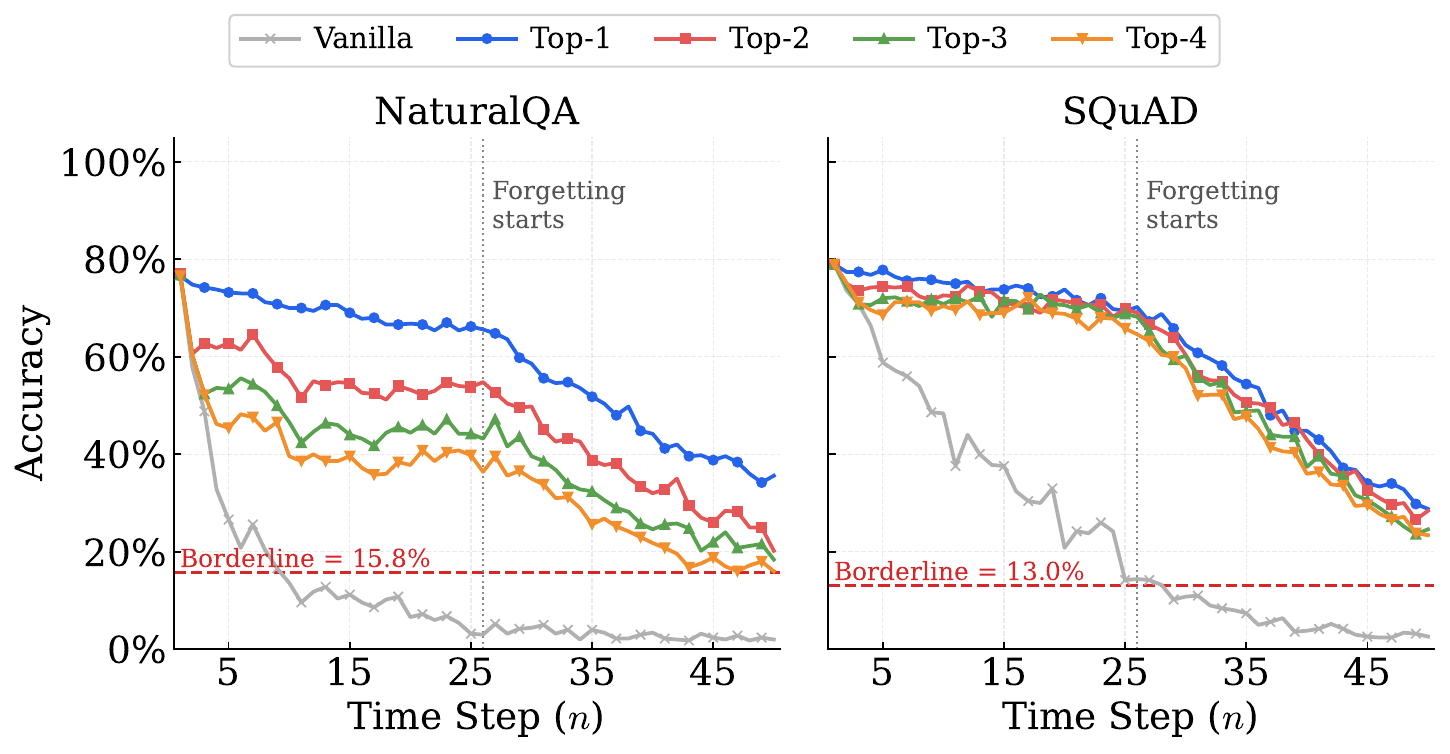}
        \caption{Qwen2.5-7B-Instruct.}
        \label{fig:retention_qwen}
    \end{subfigure}
    \hfill
    \begin{subfigure}[b]{\linewidth}
        \centering
        \includegraphics[width=\textwidth]{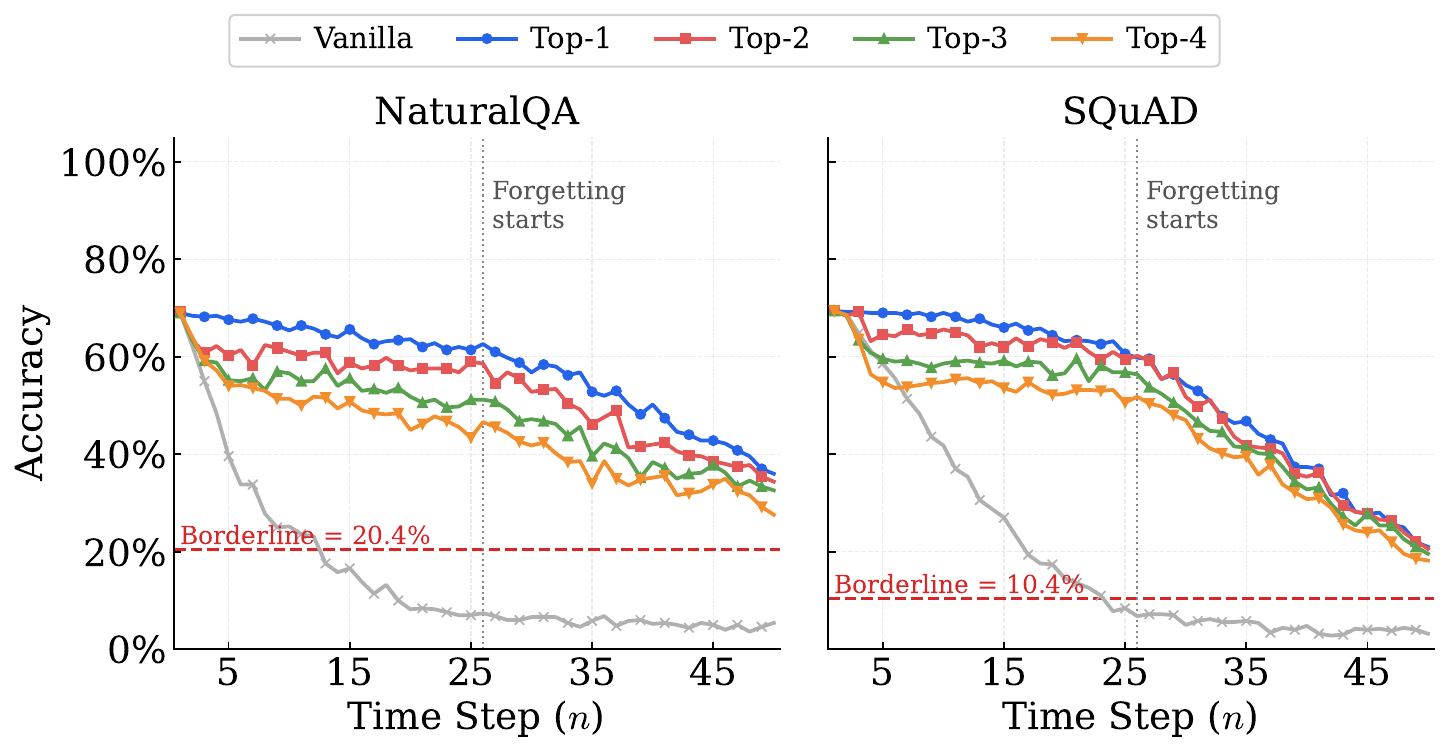}
        \caption{Mistral-7B-Instruct-v0.3.}
        \label{fig:retention_mistral}
    \end{subfigure}
    \hfill
    \begin{subfigure}[b]{\linewidth}
        \centering
        \includegraphics[width=\textwidth]{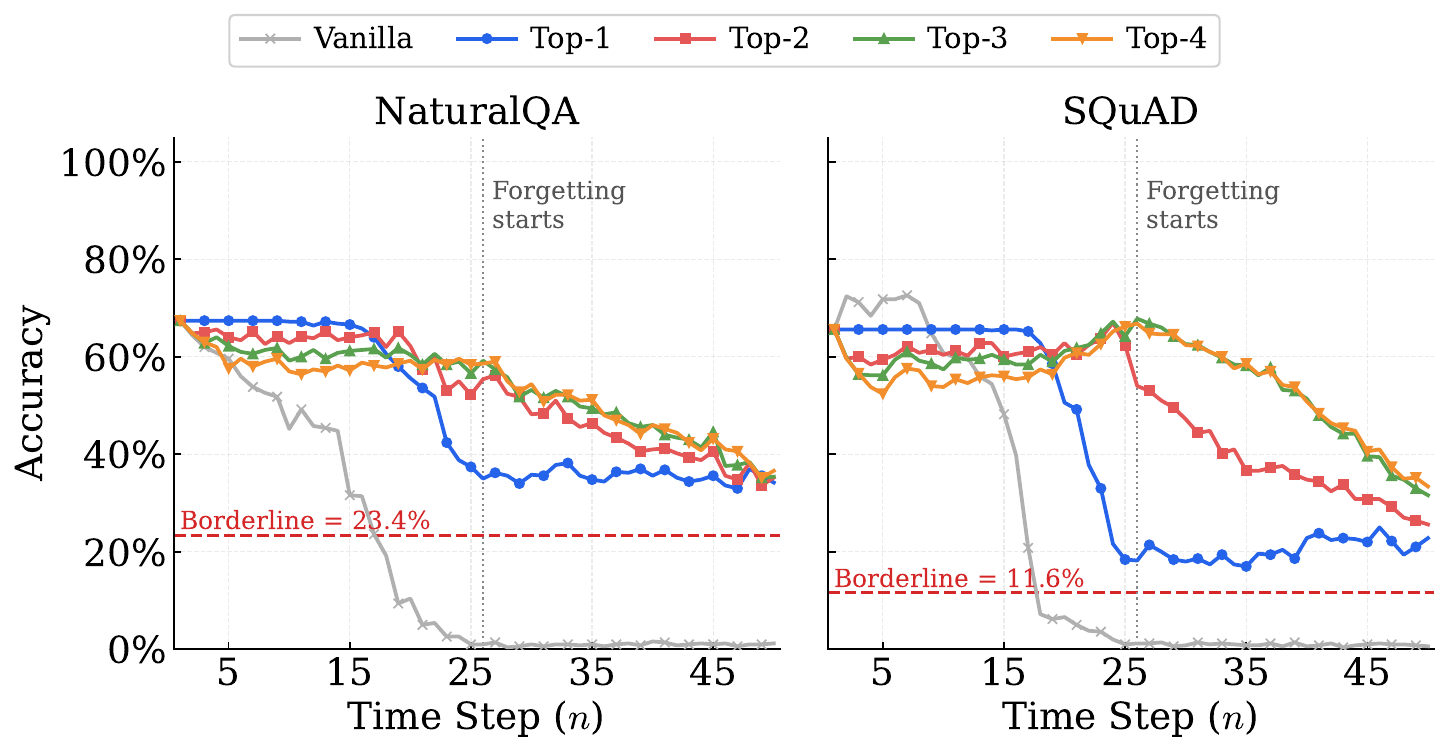}
        \caption{Gemma-2-9b-it.}
        \label{fig:retention_gemma}
    \end{subfigure}
  
    \caption{50-step knowledge retention comparing
MemDefrag (with Top-1, 2, 3 or 4 filtering) vs. vanilla latent memory on NaturalQA and SQuAD.}
    \label{fig:three_model_retention}
\end{figure}

\begin{table*}[!t]
\centering
\resizebox{\textwidth}{!}{
\begin{tabular}{l|c|ccc|ccc}
\toprule
\multirow{2}{*}{Model} & \multirow{2}{*}{Layers} & \multicolumn{3}{c|}{NaturalQA} & \multicolumn{3}{c}{SQuAD} \\
\cmidrule(lr){3-5} \cmidrule(lr){6-8}
 & & Layer & $\overline{\text{Rank}}\downarrow$ & Top-2$\uparrow$ & Layer & $\overline{\text{Rank}}\downarrow$ & Top-2$\uparrow$ \\
\midrule
Llama-3.1-8B-Instruct     & 32 & 13 (last-token) & 1.66 & 94.7 & 13 (all-token)  & 1.90 & 86.1 \\
Qwen2.5-7B-Instruct       & 28 & 14 (all-token)  & 2.02 & 90.3 & 14 (all-token)  & 1.45 & 95.2 \\
Mistral-7B-Instruct-v0.3  & 32 & 15 (all-token)  & 1.70 & 93.4 & 15 (all-token)  & 1.22 & 96.6 \\
Gemma-2-9b-it             & 42 & 15 (last-token) & 1.24 & 97.8 & 15 (all-token)  & 1.27 & 99.3 \\
\bottomrule
\end{tabular}}
\caption{Top-ranked tracer layer (i.e., the layer with the lowest $\overline{\text{Rank}}$) for each model on NaturalQA and SQuAD. For each (model, dataset) pair, we report the better of the two attention computation strategies, with the chosen strategy indicated in parentheses after the layer index. Top-2 accuracies (\%) of tracing are also reported.}
\label{tab:top_tracing_layer_summary}
\end{table*}

\subsection{Generality across Models}
\label{app:Knowledge Retention Experiments Based on Various Models}

In addition to Llama-3.1-8B-Instruct, we study the effects of MemDefrag in  Qwen2.5-7B-Instruct~\cite{qwen2.5}, Mistral-7B-Instruct-v0.3~\cite{jiang2023mistral7b}, Gemma-2-9b-it~\cite{gemma_2024}.
Table~\ref{tab:top_tracing_layer_summary} reports the Top-ranked tracer layer for each model on NaturalQA and SQuAD under last-token or all-token attention computation strategy. Two consistent patterns emerge:

(1) The Top-ranked tracer layer is consistently the middle transformer layers, concentrating within layers 13--15 for the four models. 

(2) At these top layers, \textbf{Top-2 tracing accuracy generally exceeds 90\%} across both datasets (reaching up to 99.3\% on Gemma-2-9b-it and SQuAD). This shows that the tracing signal is not an artifact of a single model, but a general phenomenon across various instruction-tuned LLMs.

Detailed results of the tracing capabilities of the four models are shown in Tables~\ref{tab:retrieval_layers_combined_Llama-3.1-8B-Instruct},~\ref{tab:retrieval_layers_combined_Qwen2.5-7B-Instruct},~\ref{tab:retrieval_layers_combined_Mistral-7B-Instruct-v0.3}, and~\ref{tab:retrieval_layers_combined_Gemma-2-9b-it}.

Further, Figure~\ref{fig:three_model_retention} presents the results of 50-step knowledge retention based on the models: Qwen2.5-7B-Instruct, Mistral-7B-Instruct-v0.3, and Gemma-2-9b-it.
The attention computation strategy for each (model, dataset) pair follows the configurations reported shown in Table~\ref{tab:top_tracing_layer_summary}. We make the following observations.

(1) The performance achieved by MemDefrag with Top-1 to Top-4 filtering substantially outperforms that of vanilla latent memory. This suggests that MemDefrag is broadly effective and can be generalized to various language models.

(2) Similar to experiment results (Figure~\ref{fig:retention_comparison} and Table~\ref{tab:retention_comparison}) of knowledge retention based on Llama-3.1-8B-Instruct, Top-1 or Top-2 filtering generally achieves the best performance across all time steps on  Qwen2.5-7B-Instruct and Mistral-7B-Instruct-v0.3.

(3) However, for Gemma-2-9b-it, the performance of Top-1 filtering drops significantly once the time step exceeds 20 and Top-1 cannot serve as the optimum choice any longer.
These results indicate that, in some cases, as the time step increases, adaptive Top-$K$ may have the potential to serve as an effective strategy for maintaining optimal performance across different time steps.

\paragraph{Plug-and-play justification of the tracer layer.}
A natural concern is whether selecting the tracer layer requires running an investigation for every new backbone. We argue that this is \emph{not} the case. As summarized in Tables~\ref{tab:retrieval_layers_combined_Llama-3.1-8B-Instruct},~\ref{tab:retrieval_layers_combined_Qwen2.5-7B-Instruct},~\ref{tab:retrieval_layers_combined_Mistral-7B-Instruct-v0.3}, and~\ref{tab:retrieval_layers_combined_Gemma-2-9b-it}, across all four instruction-tuned LLMs we examined---Llama-3.1-8B-Instruct (layer~13/32), Qwen2.5-7B-Instruct (layer~14/28), Mistral-7B-Instruct-v0.3 (layer~15/32), and Gemma-2-9b-it (layer~15/42)---the top-ranked tracer layer consistently falls within the band $[L/3,\, L/2]$, where $L$ denotes the total number of transformer layers. This band is narrow in absolute terms (6--8 candidate layers per model in our setup). In practice, this means that for a new backbone a user does not need to repeat the full investigation study of exploring the best tracing layer: A lightweight sweep over the $\lfloor L/3 \rfloor$--$\lceil L/2 \rceil$ band is sufficient to identify a high-quality tracer layer. This keeps MemDefrag genuinely \textbf{plug-and-play} in deployment.

\begin{figure}[!t]
    \centering
    \includegraphics[width=\linewidth]{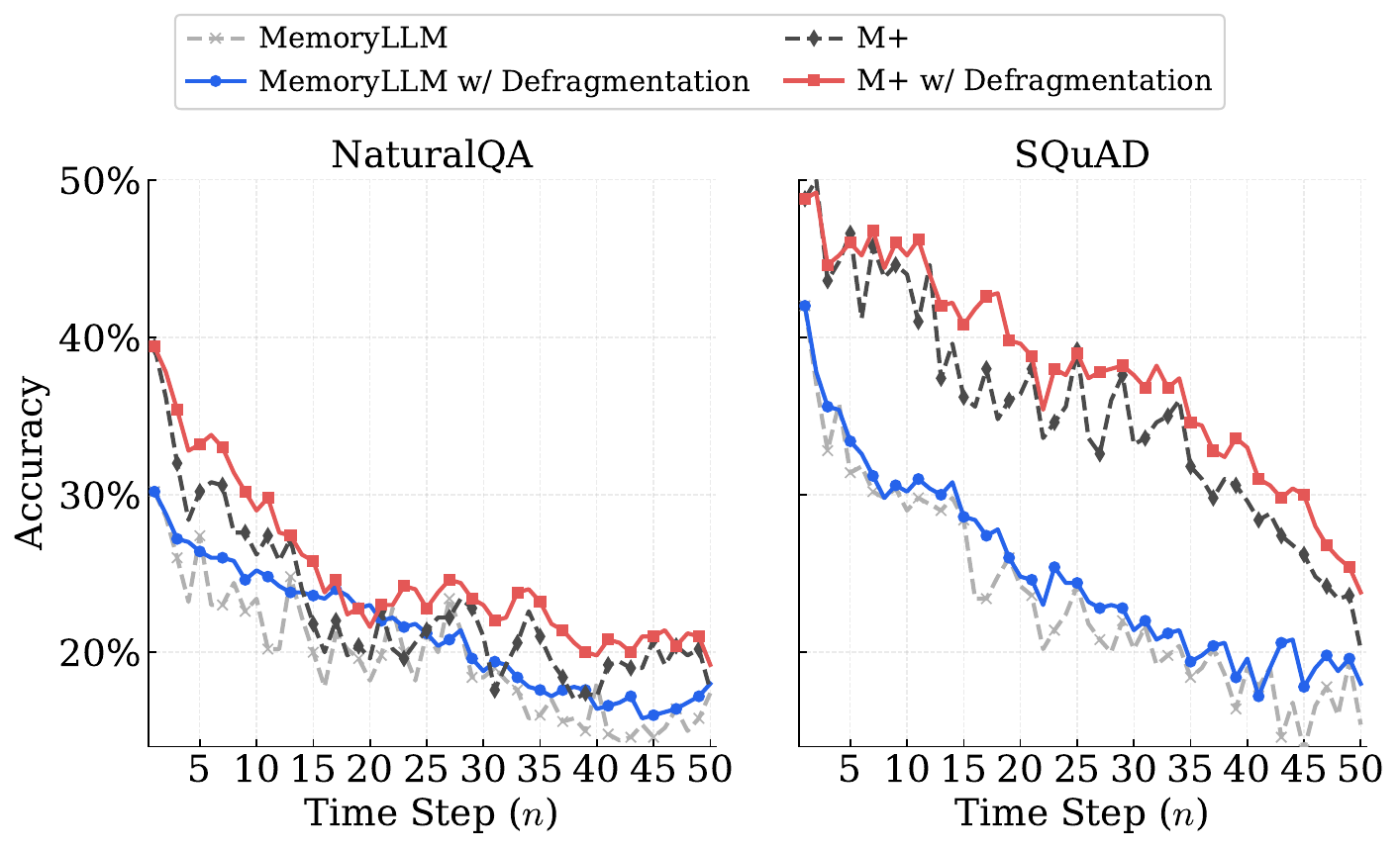}
    \caption{50-step knowledge retention on NaturalQA and SQuAD, comparing the presence and absence of defragmentation on MemoryLLM and M+.}
    \label{fig:defragmentation_memoryllm_mplus}
\end{figure}

\begin{table}[t]
\centering
\resizebox{\linewidth}{!}{%
\begin{tabular}{l ccccc ccccc}
\toprule
& \multicolumn{5}{c}{\textbf{NaturalQA}} & \multicolumn{5}{c}{\textbf{SQuAD}} \\
\cmidrule(lr){2-6} \cmidrule(lr){7-11}
\textbf{Method} & $n$=10 & $n$=20 & $n$=30 & $n$=40 & $n$=50 & $n$=10 & $n$=20 & $n$=30 & $n$=40 & $n$=50 \\
\hline
\rowcolor{blue!8} \multicolumn{11}{c}{\textit{MemoryLLM}} \\ \hline
MemoryLLM                  & 23.4 & 18.2 & 18.4 & \textbf{18.0} & 17.4 & 29.0 & 24.2 & 20.2 & 19.0 & 15.4 \\
\,\, w/ Defragmentation     & \textbf{25.2} & \textbf{23.0} & \textbf{18.8} & 16.4 & \textbf{18.0} & \textbf{30.2} & \textbf{24.8} & \textbf{21.4} & \textbf{19.6} & \textbf{18.0} \\
\hline
\rowcolor{blue!8} \multicolumn{11}{c}{\textit{M+}} \\ \hline
M+                         & 26.2 & 19.6 & 21.0 & 17.2 & 17.6 & 44.0 & 36.4 & 33.2 & 29.6 & 20.2 \\
\,\, w/ Defragmentation     & \textbf{29.0} & \textbf{21.6} & \textbf{23.0} & \textbf{19.8} & \textbf{19.2} & \textbf{45.2} & \textbf{39.6} & \textbf{37.6} & \textbf{33.0} & \textbf{23.8} \\
\bottomrule
\end{tabular}%
}
\caption{QA accuracy (\%) of MemoryLLM and M+ with and without defragmentation at selected time steps on NaturalQA and SQuAD.
\textbf{Bold} indicates the best result in each column.}
\label{tab:defrag_memoryllm_mplus}
\end{table}

\subsection{Compatibility with Latent-Memory Variants}
\label{app:Compatibility with Latent-Memory Variants}
We study the compatibility of our inference-time memory defragmentation (Section~\ref{sec:Inference-Time Memory Defragmentation}) with the other two latent-memory variants: MemoryLLM (Memoryllm-8b) and M+ (M+-8b). Considering that these two models have been with a fixed-length prefix latent memory, the defragmentation procedure in this section adopts only the ranking and reordering strategies, without Top-$K$ filtering. The experiment setup follows that of the knowledge retention experiments in Section~\ref{sec:Knowledge Retention Experiments}.

Figure~\ref{fig:defragmentation_memoryllm_mplus} and Table~\ref{tab:defrag_memoryllm_mplus} show the 50-step accuracy results under the presence or absence of defragmentation for MemoryLLM and M+. Last-token attention computation strategy based on layer 13 (the tracing layer) is used in defragmentation for both MemoryLLM and M+ and both datasets.
Two observations stand out.

\textbf{(1) Defragmentation consistently improves both backbones.} Across nearly all checkpoints on both NaturalQA and SQuAD, applying our defragmentation procedure yields a clear and stable accuracy gain over the corresponding vanilla baseline.
This indicates that explicitly reorganizing memory fragments makes the stored knowledge substantially more retrievable. 
Further, as MemoryLLM and M+ compress knowledge fragments by their own models, they support the compatibility of our MemDefrag with model-level compression.

\textbf{(2) Random forgetting still drives a steady performance decay.} Despite these gains, both MemoryLLM and M+ rely on a \emph{random} forgetting strategy. As a result, even after defragmentation, accuracy continues to exhibit an overall downward trend as $n$ grows
on both datasets.

\begin{figure}[!t]
    \centering
    \includegraphics[width=\linewidth]{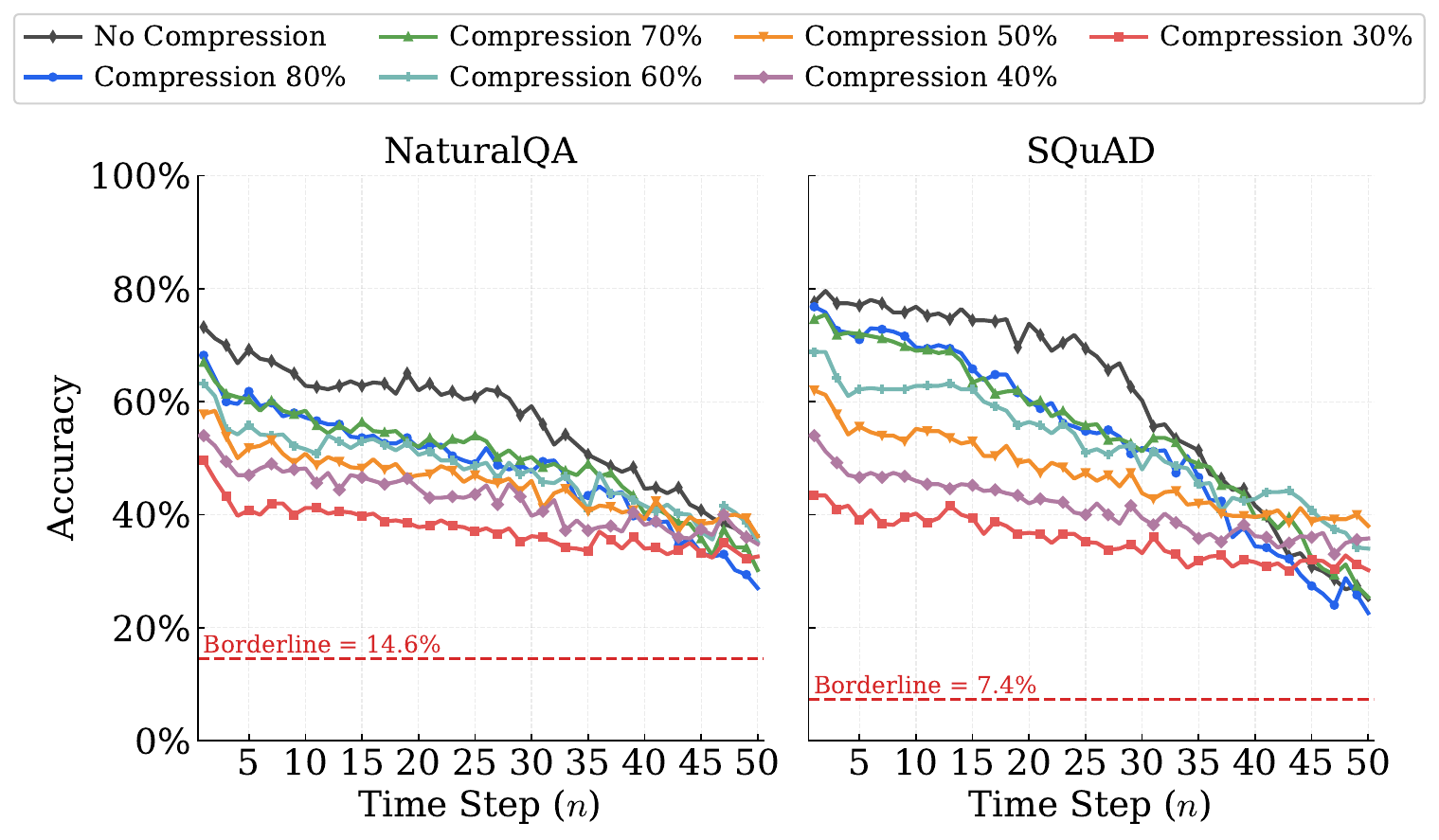}
    \caption{50-step knowledge retention on NaturalQA and SQuAD under different compression ratios.}
    \label{fig:retention_different_compression}
\end{figure}

\begin{table}[t]
\centering
\resizebox{\linewidth}{!}{%
\begin{tabular}{l ccccc ccccc}
\toprule
& \multicolumn{5}{c}{\textbf{NaturalQA}} & \multicolumn{5}{c}{\textbf{SQuAD}} \\
\cmidrule(lr){2-6} \cmidrule(lr){7-11}
\textbf{Ratio} & $n$=10 & $n$=20 & $n$=30 & $n$=40 & $n$=50 & $n$=10 & $n$=20 & $n$=30 & $n$=40 & $n$=50 \\
\hline
\rowcolor{gray!25} 100\% & \textbf{62.8} & \textbf{62.0} & \textbf{59.2} & \textbf{44.6} & \textbf{36.2} & \textbf{76.8} & \textbf{73.8} & \textbf{60.2} & 41.6          & 25.0          \\
80\%           & 57.2          & 51.8          & 47.6          & 39.6          & 27.0          & 69.6          & 60.2          & 51.8          & 34.4          & 22.6          \\
70\%           & 58.4          & 52.0          & 50.2          & 40.2          & 30.2          & 69.0          & 59.4          & 51.2          & 39.6          & 25.4          \\
60\%           & 51.6          & 50.6          & 48.0          & 41.6          & 35.2          & 62.8          & 56.4          & 48.0          & \textbf{42.8} & 34.0          \\
50\%           & 50.8          & 47.0          & 46.0          & 39.2          & \textbf{36.2} & 55.2          & 49.6          & 43.8          & 39.6          & \textbf{38.0} \\
40\%           & 48.2          & 44.6          & 39.8          & 38.2          & 34.8          & 46.0          & 42.0          & 39.4          & 36.2          & 35.8          \\
30\%           & 41.2          & 37.8          & 36.2          & 34.0          & 32.6          & 40.2          & 36.8          & 33.2          & 31.6          & 30.2          \\
\bottomrule
\end{tabular}%
}
\caption{QA accuracy (\%) under different compression ratios at selected time steps on NaturalQA and SQuAD. 100\% ratio denotes that no compression is used. 
\textbf{Bold} indicates the best result in each column.}
\label{tab:compression_comparison}
\end{table}

\subsection{Compatibility with Prompt Compression}
\label{app:Compatibility with Prompt Compression}
Unlike MemoryLLM or M+ that construct memory fragments whose lengths are 256 in a fixed manner, our MemDefrag can take advantage of approaches of prompt compression to save the memory storage. In this section, we study the effects of LLMLingua-2~\cite{pan-etal-2024-llmlingua} on MemDefrag. The experiment setup follows that of knowledge retention experiments (Section~\ref{sec:Knowledge Retention Experiments}). Specifically, the filtering strategy is Top-3 and the adopted model is Llama-3.1-8B-Instruct. Each knowledge fragment is compressed by LLMLingua-2 with a compression ratio before being formed into its memory fragment.

Figure~\ref{fig:retention_different_compression} and Table~\ref{tab:compression_comparison} show the 50-step accuracy results under different compression ratios (from 30\% to 100\%, i.e., no compression). 
The results show that \textbf{MemDefrag effectively leverages prompt compression to improve long-term knowledge retention}. Compared with the no-compression setting, moderate compression ratios maintain competitive accuracies in early steps while achieving better performance as the step number increases. 
This improvement can be attributed to the shorter length of the compressed knowledge and the ability of the adopted compression technique to preserve informative tokens. The memory grows more slowly over time, which delays the triggering of the forgetting mechanism. This explains why compressed settings become particularly advantageous at later time steps, especially when $n \geq 40$, where uncompressed knowledge (and memory) fragments have suffered from earlier forgetting and performance degradation.

\section{Scalability to Larger Backbones}
\label{sec:larger_backbones}

To evaluate the scalability of MemDefrag, we conducted the 50-step
knowledge-retention experiment described in
Section~\ref{sec:Knowledge Retention Experiments} on NaturalQA using four larger backbones: Qwen2.5-32B-Instruct~\cite{qwen2.5}, Mistral-Small-24B~\cite{mistralsmall24b2025}, Gemma-2-27b-it~\cite{gemma_2024}, and Yi-1.5-34B~\cite{ai2025yiopenfoundationmodels}.
For each backbone, we performed the same one-time lightweight sweep over
the middle-layer band described in Appendix~\ref{app:Knowledge Retention Experiments Based on Various Models}
to identify the layer exhibiting the strongest tracing capability.
We evaluated 500 groups for each configuration.
Table~\ref{tab:larger_backbones} reports the accuracy at selected
memory-update steps.

\begin{table}[t]
    \centering
    \small
    \setlength{\tabcolsep}{3.2pt}
    \resizebox{\columnwidth}{!}{%
    \begin{tabular}{@{}llccccc@{}}
        \toprule
        Model & Method
        & $n{=}10$ & $n{=}20$ & $n{=}30$ & $n{=}40$ & $n{=}50$ \\
        \midrule
        Qwen2.5-32B-Instruct
        & Vanilla & 32.0 & 7.2 & 4.8 & 3.2 & 4.4 \\
        & Top-1 & \textbf{73.6} & \textbf{68.2} & \textbf{66.4}
                & \textbf{49.2} & \textbf{35.0} \\
        & Top-2 & 70.6 & 63.0 & 47.8 & 37.8 & 23.8 \\
        \midrule
        Mistral-Small-24B
        & Vanilla & 31.4 & 9.6 & 9.2 & 8.0 & 4.0 \\
        & Top-1 & \textbf{73.0} & \textbf{69.4} & \textbf{56.8}
                & \textbf{57.8} & \textbf{49.2} \\
        & Top-2 & 67.2 & 63.6 & 54.8 & 50.0 & 38.0 \\
        \midrule
        Gemma-2-27b-it
        & Vanilla & 52.2 & 27.4 & 34.0 & 21.8 & 25.6 \\
        & Top-1 & \textbf{65.4} & \textbf{63.2} & \textbf{57.6}
                & \textbf{55.0} & 43.4 \\
        & Top-2 & 60.6 & 62.0 & 54.2 & 52.4 & \textbf{46.8} \\
        \midrule
        Yi-1.5-34B
        & Vanilla & 22.8 & 6.4 & 1.6 & 1.4 & 0.2 \\
        & Top-1 & \textbf{74.4} & \textbf{74.0} & \textbf{71.6}
                & \textbf{47.8} & \textbf{35.2} \\
        & Top-2 & 65.8 & 68.8 & 60.6 & 45.6 & 24.4 \\
        \bottomrule
    \end{tabular}%
    }
    \caption{Knowledge-retention accuracy (\%) on NaturalQA for
    24B--34B backbones at selected memory-update steps. Bold indicates
    the best result for each model and update step.}
    \label{tab:larger_backbones}
\end{table}

The results lead to three observations.
(1) \textbf{The failure mode persists at larger scales.}
Vanilla latent memory degrades sharply on every backbone. At step 50,
its accuracy falls to 0--5\% on Qwen2.5-32B-Instruct, Mistral-Small-24B, and
Yi-1.5-34B, while Gemma-2-27b-it retains only 25.6\%. This confirms that
the degradation caused by accumulated positional distortion is not
specific to relatively small language models.
(2) \textbf{MemDefrag remains effective across model scales and
families.}
At step 50, Top-1 achieves accuracies of 35.0\%, 49.2\%, 43.4\%, and 35.2\% on the four backbones, improving over vanilla latent memory by 30.6, 45.2, 17.8, and 35.0 percentage points, respectively. 
The strong result on Yi-1.5-34B further extends the observed generality to an additional model family.
(3) \textbf{A small $K$ remains preferable.}
Top-1 produces the highest accuracy in 19 of the 20 reported
model--step combinations. The only exception is Gemma-2-27b-it at step
50, where Top-2 obtains 46.8\% compared with 43.4\% for Top-1. These results are consistent with the recall--attention-density trade-off discussed in Section~\ref{sec: Ablation Study of Different K}: retaining a small number of
highly ranked fragments generally provides the strongest knowledge
retention.

\begin{table*}[t]
\centering
\begin{tabular}{cl}
\toprule
\textbf{Notation} & \textbf{Description} \\
\hline
\rowcolor{blue!8} \multicolumn{2}{c}{\textit{Model \& Memory}} \\ \hline
$\mathcal{M}_{\theta,\phi}$ & Language model with memory parameters $\theta$ and model parameters $\phi$ \\
$\phi = \{\phi^l\}_{l=1}^{L}$ & Model parameters ($L$ transformer layers) \\
$\theta = \{\theta^l\}_{l=1}^{L}$ & Memory parameters (per-layer hidden vectors) \\
$d$ & Hidden dimension \\
$L$ & Number of transformer layers \\
\hline
\rowcolor{blue!8} \multicolumn{2}{c}{\textit{Knowledge \& Memory Formation}} \\ \hline
$x_i$ & Textual knowledge fragment received at time step $i$ \\
$|x_i|$ & Number of tokens in knowledge fragment $x_i$ \\
$m_i = \{m_i^l\}_{l=1}^{L}$ & Formed memory fragment for knowledge fragment $x_i$ \\
$m_i^l \in \mathbb{R}^{L_i \times d}$ & Hidden states of $x_i$ at layer $l$ \\
$n$ & Current update step, equivalently the number of stored knowledge fragments \\
\hline
\rowcolor{blue!8} \multicolumn{2}{c}{\textit{Memory Evolution \& Forgetting}} \\ \hline
$\theta_n = \{m_1, \dots, m_n\}$ & Accumulated memory after storing $n$ knowledge fragments \\
$N_n$ & Total number of hidden states in $\theta_n$, i.e., $\sum_{i=1}^{n} L_i$ \\
$N_{\max}$ & Maximum memory capacity (number of hidden states) \\
$L_i$ & Current retained length (number of hidden states) for knowledge fragment $x_i$ \\
$N_{\text{forget}}$ & Total number of hidden states to be pruned \\
$N_{\text{forget},i}$ & Forgetting quota allocated to knowledge fragment $x_i$ \\
$U_\phi(\cdot)$ & Memory update function \\
$f_{\text{form}}(\cdot)$ & Memory formation function \\
$f_{\text{forget}}(\cdot)$ & Forgetting function \\
$I(x_i^{(t)})$ & Self-information of the $t$-th token in $x_i$ \\
$D_i$ & Index set of positions to be dropped from $x_i$ \\
\hline
\rowcolor{blue!8} \multicolumn{2}{c}{\textit{Memory Utilization}} \\ \hline
$p$ & Prompt (e.g., a query) at inference time \\
$\theta_n^p$ & Prompt-specific inference-time temporary memory obtained from $\theta_n$ \\
$f_{\text{defrag}}(\cdot)$ & Defragmentation function \\
$l^*$ & Tracer layer used for attention scoring \\
$H_p^{l} \in \mathbb{R}^{|p| \times d}$ & Prompt hidden states at layer $l$ \\
$H_\theta^{l} \in \mathbb{R}^{(1+N_n) \times d}$ & Concatenated memory hidden states (including \texttt{bos}) at layer $l$ \\
$\bar{a}_t^l$ & Per-position attention score at layer $l$\\
$\rho_i^l$ & Attention density of knowledge fragment $x_i$ at layer $l$ \\
$P_i$ & Index set of positions corresponding to $x_i$ in concatenated memory \\
$\pi$ & Permutation that sorts knowledge fragments by attention density in ascending order \\
$K$ & Number of top-ranked memory fragments retained for inference \\
$N_K$ & Total number of hidden states in the retained Top-$K$ memory fragments \\
\bottomrule
\end{tabular}
\caption{Summary of key notations used in this paper.}
\label{tab:notation}
\end{table*}

\end{document}